\documentclass{article} 
\usepackage{preprint,times}

\usepackage{amsmath,amsfonts,bm}

\def\eqref#1{equation~\ref{#1}}

\def\1{\bm{1}}

\DeclareMathAlphabet{\mathsfit}{\encodingdefault}{\sfdefault}{m}{sl}
\SetMathAlphabet{\mathsfit}{bold}{\encodingdefault}{\sfdefault}{bx}{n}

\usepackage{hyperref}
\usepackage{url}

\usepackage{xspace}
\usepackage{dsfont}
\usepackage{afterpage}

\usepackage{enumitem}
\usepackage{booktabs}
\usepackage{caption} 
\usepackage{multirow} 
\usepackage{enumitem}
\usepackage{colortbl} 
\usepackage{bbm}
\usepackage{algorithm}
\usepackage{algorithmic}
\usepackage{setspace}                       
\usepackage{lipsum} 
\usepackage{subcaption} 
\usepackage{wrapfig}

\usepackage{mathtools} 
\usepackage{bm} 
\usepackage{soul} 
\usepackage{nicefrac}
\usepackage{csquotes}

\usepackage{duckuments}
\usepackage{wrapfig}

\definecolor{cornellred}{rgb}{0.7, 0.11, 0.11}
\definecolor{cadmiumgreen}{rgb}{0.0, 0.42, 0.24}
\definecolor{aliceblue}{rgb}{0.91, 0.94, 0.97}
\definecolor{darkblue}{rgb}{0.83, 0.89, 0.97}
\definecolor{Red7}{rgb}{0.941, 0.243, 0.243}
\definecolor{Green7}{RGB}{55, 178, 77}
\definecolor{Blue9}{rgb}{0.098,0.3,0.9}

\usepackage{pifont}

\sethlcolor{aliceblue}

\hypersetup{
  linkcolor = cornellred,
  citecolor  = cadmiumgreen,
  colorlinks = true,
  urlcolor = Blue9
}

\iclrfinalcopy

\title{
See Further When Clear: \\ 
Curriculum Consistency Model
}
\author{
Yunpeng Liu$^{1,2}${\quad}
Boxiao Liu$^3${\quad}
Yi Zhang$^3${\quad}
Xingzhong Hou$^{1,2}${\quad}
Guanglu Song$^3${\quad}
\\
{\space}\textbf{Yu Liu$^{3}$}{\quad}
\textbf{Haihang You$^{1,}$\thanks{corresponding author}$^\ast$}\\
$^1$Institute of Computing Technology, Chinese Academy of Sciences \,\, \\
$^2$School of Computer Science and Technology, University of Chinese Academy of Sciences\,\,\, \\
$^3$SenseTime Research\,\,\, 
}

\newcommand*{\ShowNotes}{} 

\ifdefined\ShowNotes
  \newcommand{\colornote}[3]{{\color{#1}\bf{#2: #3}\normalfont}}
\else
  \newcommand{\colornote}[3]{}
\fi

\begin{document}
\maketitle

\begin{abstract}
Significant advances have been made in the sampling efficiency of diffusion models and flow matching models, driven by Consistency Distillation (CD), which trains a student model to mimic the output of a teacher model at a later timestep. However, we found that the learning complexity of the student model varies significantly across different timesteps, leading to suboptimal performance in CD.
To address this issue, we propose the Curriculum Consistency Model (CCM), which stabilizes and balances the learning complexity across timesteps. Specifically, we regard the distillation process at each timestep as a curriculum and introduce a metric based on Peak Signal-to-Noise Ratio (PSNR) to quantify the learning complexity of this curriculum, then ensure that the curriculum maintains consistent learning complexity across different timesteps by having the teacher model iterate more steps when the noise intensity is low.
Our method achieves competitive single-step sampling Fréchet Inception Distance (FID) scores of 1.64 on CIFAR-10 and 2.18 on ImageNet 64x64.
Moreover, we have extended our method to large-scale text-to-image models and confirmed that it generalizes well to both diffusion models (Stable Diffusion XL) and flow matching models (Stable Diffusion 3). The generated samples demonstrate improved image-text alignment and semantic structure, since CCM enlarges the distillation step at large timesteps and reduces the accumulated error.

\end{abstract}

\section{Introduction}

\begin{figure*}[h]
    \centering
    \includegraphics[width=0.7\linewidth]{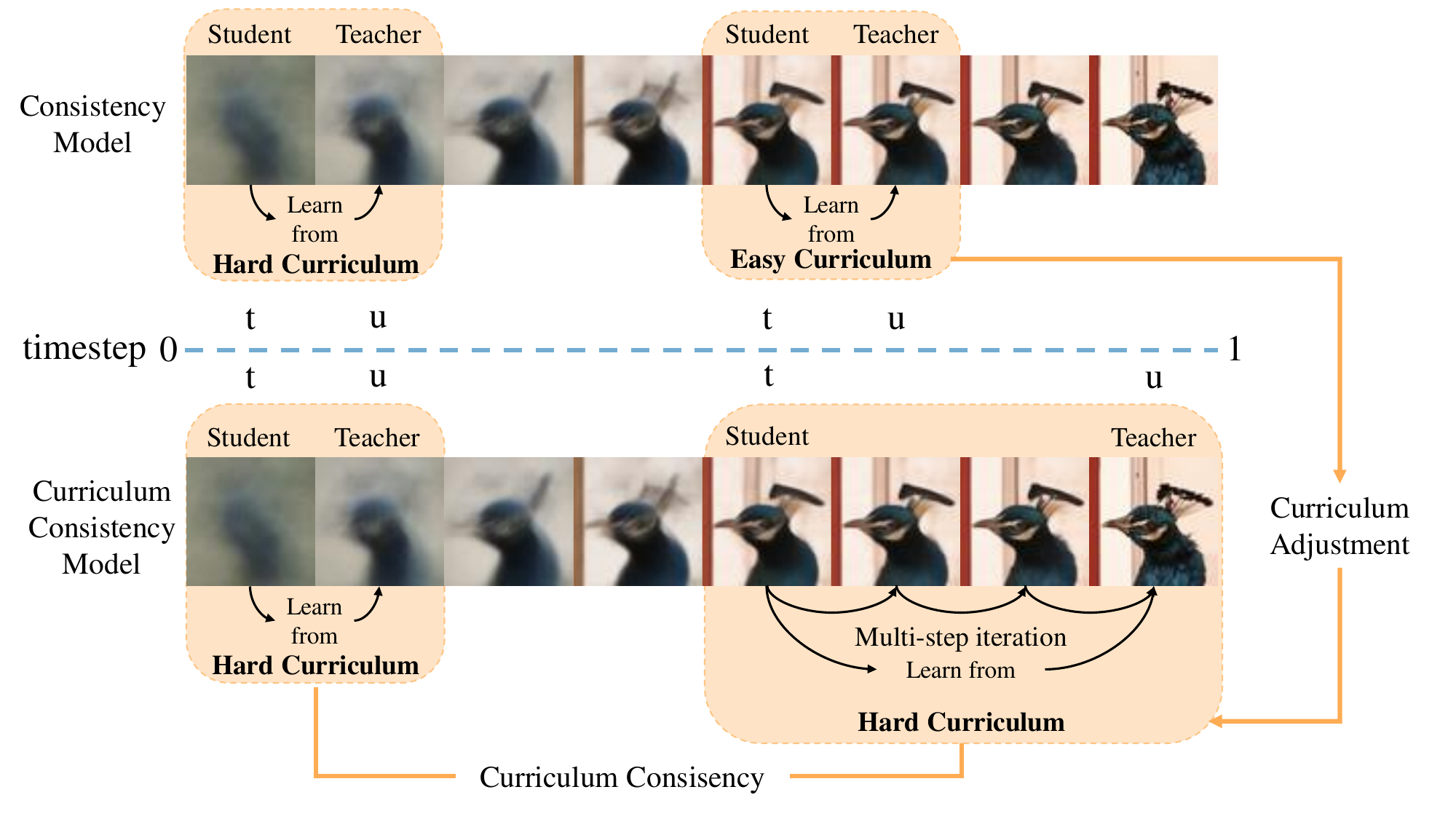}
    \caption{Comparison between Consistency Models (CM) and Curriculum Consistency Model (CCM). CM encourages the student model at timestep $t$ to learn from the teacher model at timestep $u$, but the knowledge discrepancy (curriculum difficulty) at a larger timestep is small. CCM maintains curriculum consistency by dynamically adjusting the teacher model to a more challenging timestep through multi-step iteration.}
    \label{fig:intro}
\end{figure*}

Diffusion Models (DM) and Flow Matching (FM) are two leading methods for generative image synthesis. DM \cite{ho2020denoising},\cite{songdenoising},\cite{songscore} generates samples by iteratively reversing a diffusion process, i.e., Stochastic Differential Equation (SDE), whereas FM \cite{lipman2023flow},\cite{tong2023improving} constructs explicit probability paths, known as Probability Flow Ordinary Differential Equations (PF-ODE), between noise and data, incorporating the reversed diffusion process as a special case.
Despite the ability to produce high-quality images of DM and FM, their performances in sampling efficiency are not satisfactory and often require a lot of function evaluations. With the introduction of Consistency Models (CM) \cite{song2023consistency}, the Number of Function Evaluations (NFEs) required for sampling has been significantly reduced by enforcing self-consistency. In common, as shown in Figure~\ref{fig:intro}, CM encourages the student model at timestep $t$ (where $t \in [0,1)$) to mimic the output of the teacher model at timestep $u$ (where $u \in (t, 1]$). Latent consistency models (LCM) \cite{luo2023latent} employ self-consistency in the latent space, significantly reducing computational costs and extending CM to high-resolution text-to-image syntheses, thereby promoting the widespread application of CM.

We found a critical problem in CM that differences between student and teacher outputs are highly unstable across different timesteps, resulting in inefficient training. Specifically, we regard the distillation process that student learn from teacher as a curriculum and use knowledge discrepancy to evaluate the curriculum difficulty. Easy curriculum leads to unsatisfactory generation of details ($t \rightarrow 1$) and high-level features such as semantic and structural features ($t \rightarrow 0)$. We visualize the issue in Figure \ref{fig:psnr_dataset}, where we quantify the knowledge discrepancy based on the Peak Signal-to-Noise Ratio (PSNR) between the student and teacher outputs at different timesteps. The results indicate that the knowledge discrepancy of curriculums decreases gradually as $t$ progresses from smaller values (corresponding to near-pure noise) to larger values (closer to the final image).
However, most studies \cite{song2023consistency}, \cite{luo2023latent} suffer from the instability of knowledge discrepancy, as they sample uniformly along the timesteps and use a fixed distillation step $l=u-t$ for the CM.
As a result, the student model struggles to learn effectively from easy curriculums, which affects the semantic structure and details in the diffusion process.
Recent works, iCT \cite{song2023improved} and ECM \cite{geng2024consistency}, have also tackled similar instabilities in CMs. However, their focus is on addressing error accumulation, known as the "Curse of Consistency" \cite{geng2024consistency}. iCT progressively reduces the distillation step following a power-law schedule during training, while ECM refines this reduction process to achieve a smoother transition from diffusion models to consistency models. As shown in Figure~\ref{fig:psnr_l}, decreasing the distillation step reduces knowledge discrepancy, which makes the training inefficiency more obvious.

To address these issues, we propose an adaptive training method that stabilizes and balances the knowledge discrepancy under varying noise intensities, as shown in Figure~\ref{fig:intro}. 
We first measure the Knowledge Discrepancy of the Curriculum (KDC) based on PSNR at the current timestep. Then our approach dynamically adjusts the learning targets to construct a hard curriculum with reasonable knowledge discrepancy. To ensure high-quality teacher outputs, we efficiently adopt a multi-step iterative generation strategy.

In summary, we propose the Curriculum Consistency Model (CCM) to perform the consistency distillation for the diffusion models and flow matching models. 
Our main contributions are as follows:
\begin{itemize}
    \item We identify the instability in knowledge discrepancy during consistency distillation, which significantly impacts text-to-image alignment and the generation of semantic structures in the diffusion process.
    \item We introduce a metric KDC based on PSNR to assess curriculum difficulty and design a more effective adaptive noise schedule to maintain curriculum consistency across different training samples.
    \item Our method achieves high-quality few-step generation. Specifically, we obtain one-step sampling Fréchet Inception Distance (FID) scores of 1.64 on CIFAR-10 and 2.18 on ImageNet 64x64.
    \item CCM generalizes well and has been extended to both large-scale diffusion models (Stable Diffusion XL \cite{podell2024sdxl}) and flow matching models (Stable Diffusion 3 \cite{esser2024scaling}) for high-resolution image generation. Our results show that the introduction of curriculum consistency leads to lower FID, higher CLIP scores, and significantly improved image-text alignment and semantic structure in the generated images.
\end{itemize}
\section{Related Works}

\paragraph{Diffusion Models (DM)}. 
Diffusion models have become a leading approach in high-fidelity image generation \cite{rombach2022high}, \cite{hoogeboom2023simple}. This type of model relies on Stochastic Differential Equations (SDEs) to find trajectories from noise to data. Recent work focuses on improving sample quality \cite{ho2020denoising}, optimizing density estimation \cite{songscore}, and accelerating the sampling process \cite{song2023consistency}, \citep{phung2023wavelet}.
Some studies explore the underlying mechanisms and design space of DMs \cite{karras2022elucidating}, while others scale up DMs for text-conditioned image synthesis \cite{podell2024sdxl} or improve sampling efficiency through methods in the latent space\cite{songdenoising}. 

\paragraph{Flow Matching (FM)}. 
Flow matching models learn a vector field that generates an Ordinary Differential Equation (ODE) for a desired trajectory from noise to data, without requiring computationally intensive simulations \cite{lipman2023flow}. This flexibility has led to various efforts to improve trajectory properties, particularly straightness, which enables efficient simulation with fewer steps. Methods like rectified flow \cite{liu2022flow}, \cite{liu2022rectified}, multi-sample FM \cite{pooladian2023multisample}, and minibatch OT-CFM \cite{tong2023improving} aim to straighten trajectories, but the computation costs and sample efficiency are still unsatisfied.

\paragraph{Consistency Models (CM).}
Consistency models \cite{song2023consistency} represent a new family of generative models that ensures all points along the ODE trajectory converge to the same solution, often surpassing diffusion models in performance and significantly improving the sample efficiency. 
Consistency Trajectory Model (CTM) \cite{kim2023consistency} introduces trajectory consistency and further allows unlimited traversal along the PF-ODE between arbitrary starting and ending points during the diffusion process, offering a flexible framework. 
Latent diffusion models (LCM) \cite{luo2023latent} employ consistency distillation in the latent space and extend
the models to high-resolution text-to-image synthesis.
Phased Consistency Model (PCM) \cite{pcm2024wang} identifies key limitations in LCM and addresses them by phasing the ODE trajectory and enforcing the self-consistency property on each sub-trajectory. 
iCT \cite{song2023improved} improves the training of CM by removing the EMA of the teacher, adopting Pseudo-Huber loss,  adjusting the discretization and noise schedule, etc. 
Inspired by iCT, ECM \cite{geng2024consistency} studies discretization interval deeply and proposes adaptive scaling discretization interval and continuous time scheduling schemes.
SCott \cite{liu2024scott} improves sample quality and diversity by controlling noise intensity, adopting a multi-step sampling strategy. 
sCMs \cite{lu2024simplifying} analyze and improve hyperparameter issues and discretization errors in most CMs based on discretized timesteps, enabling sCMs to train continuous-time CMs at an unprecedented scale. 

\section{Method}

\subsection{Preliminaries}

Consistency models \cite{song2023consistency} aim to simplify multiple function evaluations by directly learning an Ordinary Differential Equation (ODE) that maps any point $x$ on the ODE trajectory to the same output at the endpoint. 
Specifically, suppose that 0 means noise and 1 means image, the objective of consistency distillation is to align the neural mapping $\boldsymbol{f}_\theta$ with the true mapping $\boldsymbol{f}$ by ensuring $\boldsymbol{f}_\theta(\boldsymbol{x}_t, t, 1) \approx \boldsymbol{f}(\boldsymbol{x}_t, t, 1), \forall t \in [0,1)$.
We can train $\boldsymbol{f}_\theta$ by comparing it with the numerical solution of the pre-trained ODE solver.

\begin{equation}
    \boldsymbol{f}_{\theta}(\boldsymbol{x}_t,t,1)\approx \mathrm{Solver}(\boldsymbol{x}_t,t,1;\phi)\approx \boldsymbol{f}(\boldsymbol{x}_t,t,1)
\end{equation}

where $\phi$ means a perfect teacher model.
To simplify the training process, local consistency \cite{kim2023consistency} is often performed and formulated in Eq. \ref{eq:localcon}, which compares the student's prediction with the result obtained by solving the ODE over the interval $(t, u)$ using the teacher model, followed by mapping to timestep 1:

\begin{equation}
\label{eq:localcon}
    \boldsymbol{f}_\theta(\boldsymbol{x}_t,t,1) \approx \boldsymbol{f}_{\theta^-}(\mathrm{Solver}(\boldsymbol{x}_t,t,u;\phi),u,1)
\end{equation}

where $u$ is randomly sampled from $(t, 1)$, and $\theta^-$ denotes the exponential moving average (EMA) of the parameters, $\theta^- \leftarrow stopgrad(\mu \theta^- + (1 - \mu)\theta)$. 
Local consistency ensures that the student model effectively distills information from the teacher model over the interval $(t,u)$.
After training, the generation process begins by sampling $\boldsymbol{x}_0 \sim \mathcal{N}(0,I)$, and then directly obtaining $\boldsymbol{x}_1$ through $\boldsymbol{f}_\theta(\boldsymbol{x}_0, 0, 1)$.

\textbf{Consistency Distillation in Diffusion Models}. 
In diffusion models, the inverse of the diffusion process can be represented by a deterministic ODE which is given by \cite{songscore} :

\begin{equation}
    \mathrm{d} \boldsymbol{x}=\begin{bmatrix}-\frac12 \boldsymbol{\beta}_\sigma \boldsymbol{x}_\sigma -\frac12 \boldsymbol{\beta}_\sigma\mathbf{s}_\theta(\boldsymbol{x}_\sigma,\sigma)\end{bmatrix} \mathrm{d}\sigma
\end{equation}

where $\sigma \in [\epsilon,T]$ means noise-to-signal ratio and $\epsilon$ is a small positive value to ensure numerical stability, $\boldsymbol{\beta}$ is variance and $\mathbf{s}_\theta$ is score function. Note that the noise-to-signal ratio can be transfered into timestep through $\sigma = \frac{1-t}{t}$, so the neural mapping $\boldsymbol{f}_\theta$ in diffusion models can be described by $\sigma$: $\boldsymbol{f}_{\theta}(\boldsymbol{x}_\sigma, \sigma, \epsilon) \approx \boldsymbol{x}_\epsilon$.

A practical solution is to enforce consistency between two adjacent points (timesteps) on the ODE trajectory. By discretizing the interval $[\epsilon, T]$ into $N$ steps, $\sigma_i = \left( \epsilon^{1/\rho} + \frac{i-1}{N-1}(T^{1/\rho} - \epsilon^{1/\rho}) \right)^\rho$ \cite{karras2022elucidating}, we can approximate $\hat{\boldsymbol{x}}_\phi(\sigma_n)$ using Euler's method, and the resulting loss function is:

\begin{equation}
\label{eq:cd_loss}
\begin{split}
    \mathcal{L}_{\mathrm{CD}}^N(\theta,\theta^-; \phi) = 
    \mathbb{E}_{n\sim\mathcal{U}[1,N-1]} \Big[ \boldsymbol{\lambda}(\sigma_n) 
    d\left(\boldsymbol{f}_\theta(\boldsymbol{x}_{\sigma_{n+1}},\sigma_{n+1}, \epsilon), \right. 
    \left. \boldsymbol{f}_{\theta^-}(\hat{\boldsymbol{x}}_{\phi,\sigma_n},\sigma_n, \epsilon)\right) \Big]
\end{split}
\end{equation}

where $\lambda(\sigma_n) = 1$ and $d(\cdot, \cdot)$ is a distance metrics. 

\textbf{Consistency Distillation in Flow Matching}.
Continuous Normalizing Flow (CNF) $\boldsymbol{\psi}_t(\boldsymbol{x})$ transforms a probability density from $\boldsymbol{p}_0$ to $\boldsymbol{p}_1$ \cite{chen2018neural}, which is a time-dependent diffeomorphic map induced by vector field $\boldsymbol{u}_t(x)$, can be derived using the ODE:

\begin{equation}
    \mathrm{d} \boldsymbol{\psi}_t(\boldsymbol{x})=\boldsymbol{u}_t(\boldsymbol{\psi}_t(\boldsymbol{x})) \mathrm{d}t, \quad \boldsymbol{\psi}_0(\boldsymbol{x}_0)=\boldsymbol{x}_0
\end{equation}

Conditional Flow Matching (CFM) \cite{lipman2023flow} is a simplified simulation-free framework for training CNFs by regressing onto a target vector field $\boldsymbol{u}_t(\boldsymbol{x})$. A specific choice of the ODE trajectory is the optimal transport displacement interpolant and the corresponding trajectory points $\boldsymbol{x}_t =\boldsymbol{\psi}_t(\boldsymbol{x}_0|\boldsymbol{x}_1)=(1-t)\boldsymbol{x}_0+t\boldsymbol{x}_1$. Then we can implement consistency distillation based on Eq. \ref{eq:localcon}. 
Specific consistency distillation in flow matching has not been extensively studied, which has also been deeply explored in this paper.

\subsection{Problem Analysis}
\label{sec:3}
In generative models based on denoising, the varying levels of noise in the input can lead to different signal-to-noise ratios (SNR) during the denoising process, as discussed in \cite{karras2022elucidating,hang2023efficient}. Consequently, at different training timesteps, the difficulty that generative models learn varies, which in turn affects the model's convergence rate and the quality of the generated results. The core of the knowledge discrepancy lies in the magnitude of the difference between the model's predicted results and the ground truth. Inspired by this phenomenon, we conducted an in-depth examination of the knowledge discrepancy during the consistency model learning process by comparing the outputs of the student model with those of the teacher model. 

In this article, we regard the distillation information over the interval $(t,u)$ as a curriculum and propose a metric based on the Peak Signal-to-Noise Ratio (PSNR) to access knowledge discrepancy of the curriculum, as PSNR is widely used to measure the difference between a denoised image and its original counterpart. Specifically, according to Eq. \ref{eq:localcon}, given the outputs of the student model, $\boldsymbol{x}_\mathrm{est} = \boldsymbol{f}_\theta(\boldsymbol{x}_t,t,1)$, and those of the teacher model, $\boldsymbol{x}_\mathrm{target} = \boldsymbol{f}_{\theta^-}(\mathrm{Solver}(\boldsymbol{x}_t,t,u;\phi),u,1)$, Knowledge Discrepancy of the Curriculum (KDC) over the interval $(t,u)$ is defined as $\mathrm{KDC}_t^u$ and calculated using the following formula: 
\begin{equation}
\label{eq:KDC}
\begin{split}
    \mathrm{KDC}_t^u & = 100 - \mathrm{PSNR}(\boldsymbol{x}_{\mathrm{est}}, \boldsymbol{x}_{\mathrm{target}}) \\
    & = 100 - 10 \log_{10} (\frac{(2^n-1)^2}{\text{MSE}(\boldsymbol{f}_\theta(\boldsymbol{x}_t,t,1), \boldsymbol{f}_{\theta^-}(\mathrm{Solver}(\boldsymbol{x}_t,t,u;\phi),u,1)})
\end{split}
\end{equation}

$n$ represents the bit depth of the image. A large KDC means large difference between $\boldsymbol{x}_{\mathrm{est}}$ and $\boldsymbol{x}_{\mathrm{target}}$, and vice versa.

\begin{wrapfigure}{r}{0.38\textwidth} 
    \centering
    \vspace{-15pt}
    \includegraphics[width=\linewidth]{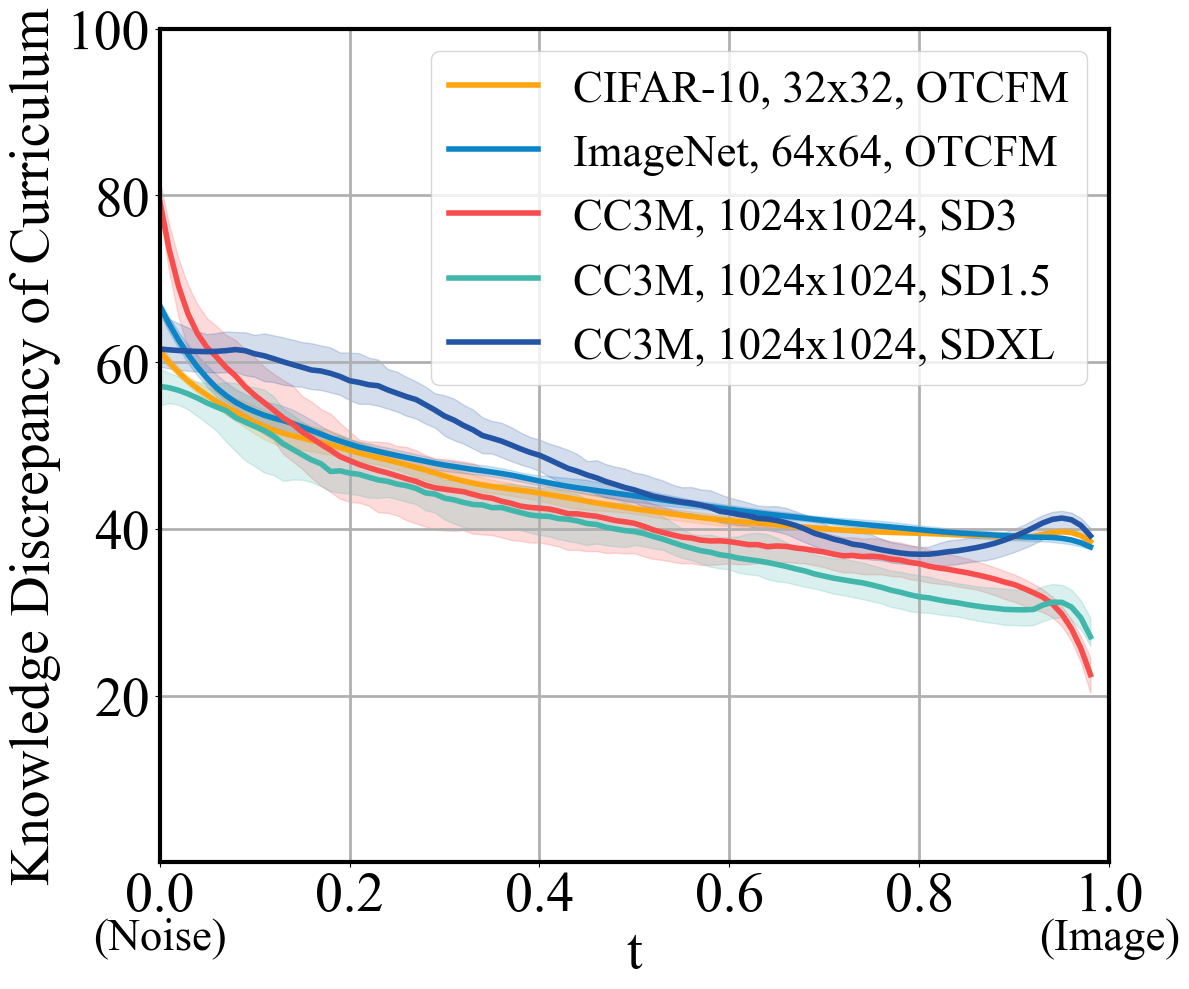}
    \caption{Knowledge Discrepancy Investigation: Analysis of the KDC over $(t,u)$ across different timesteps on various datasets for both flow matching models and diffusion models.}
    \label{fig:psnr_dataset}
    \vspace{-5pt}
\end{wrapfigure}
We conducted measurements on both diffusion models (SD 1.5~\cite{rombach2022high}, SDXL~\cite{podell2024sdxl}) and flow matching models (SD3~\cite{esser2024scaling}, OTCFM~\cite{tong2023improving}) and select 3 classic datasets (CIFAR-10, ImageNet, and CC3M) covering both low and high resolutions (32x32, 64x64, and 1024x1024) to ensure reliability and robustness. 
The mean and variance of KDC between the student and teacher model outputs on $t$ are shown in Figure \ref{fig:psnr_dataset}. 
KDC shows similar trends and close values across different datasets and models, demonstrating that it is a stable and intuitive indicator for measuring knowledge discrepancy during consistency distillation.
We observe that the KDC value consistently decreases as $t$ progresses from 0 to 1, indicating a gradual reduction in the knowledge discrepancy of curriculums. This aligns with our intuition: when $t$ is near 0, the KDC is typically around 60, as the input is heavily mixed with noise, leading to a large knowledge discrepancy. At this stage, the model is prone to confusion, causing instability and slow convergence. Conversely, when $t$ approaches 1, the KDC is usually less than 40, indicating that the knowledge discrepancy is too small, resulting in reduced learning efficiency. We argue that this instability and inefficiency hinder the overall learning process of the CM. 

\begin{wrapfigure}{r}{0.38\textwidth} 
    \centering
    \vspace{-15pt}
    \includegraphics[width=\linewidth]{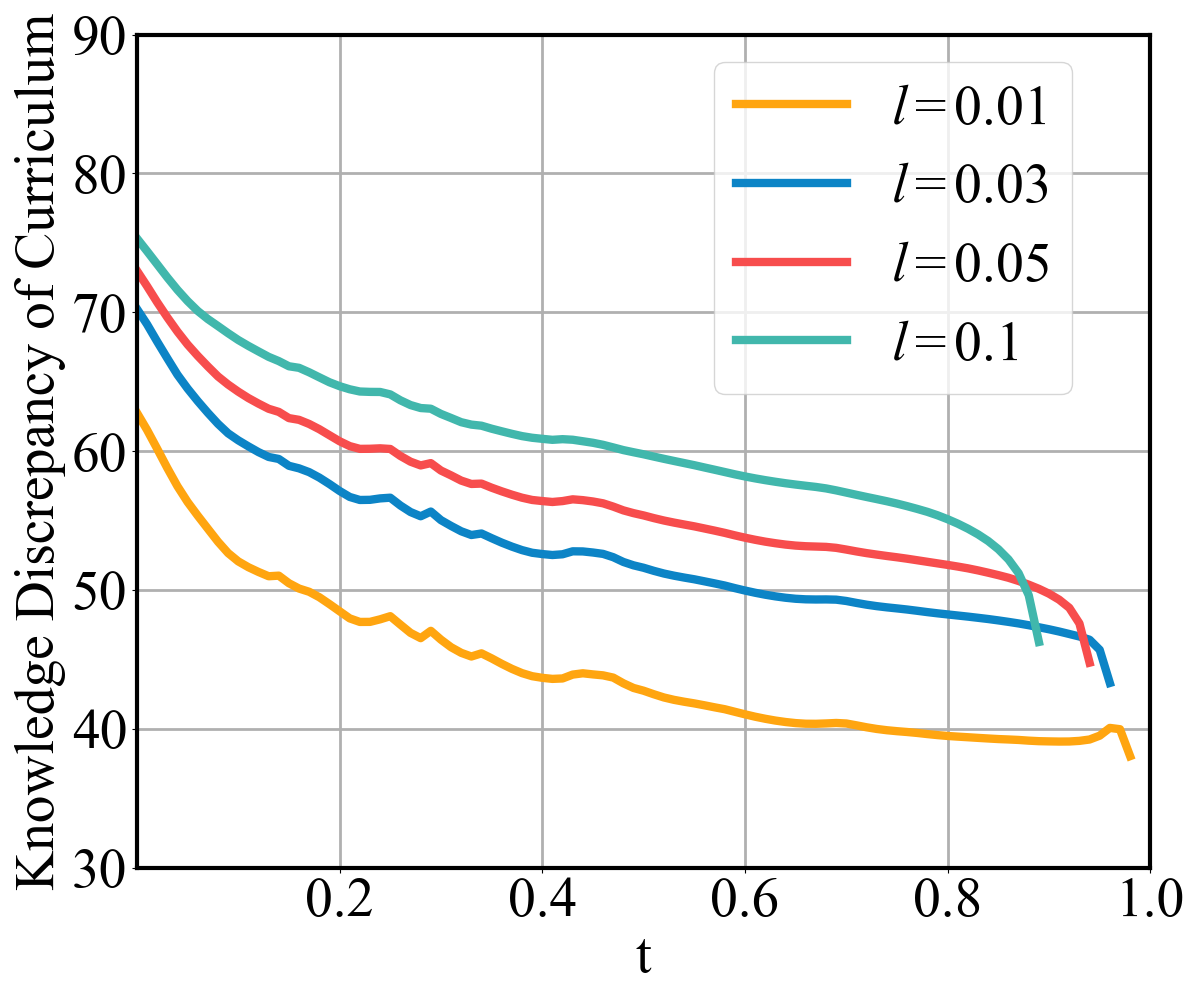}
    \caption{The relationship of KDC with different distillation steps $l$.}
    \vspace{-25pt}
    \label{fig:psnr_l}
\end{wrapfigure}
We further explored the effect of distillation step \( l=u-t \) in CM, and the results are presented in Figure~\ref{fig:psnr_l}. 
It can be observed that KDC decreases as 
$l$ decreases. Consequently, in iCT \cite{song2023improved} and ECM \cite{geng2024consistency}, where $l$ reduces over training iterations, the progressively smaller differences between student and teacher model outputs are more prone to cause inefficient learning.

Can we mitigate this imbalance in knowledge discrepancy to enhance the effectiveness of CM learning? In this paper, we attempt to present a feasible solution by proposing an adaptive method named the Curriculum Consistency Model (CCM) which will be elaborated in the following section.

\subsection{Curriculum Consistency Model}
Our goal is to design an algorithm that ensures a stable and balanced knowledge discrepancy for the model at different timesteps (i.e., under different noise intensities) and various training iterations. To achieve this, we should \textbf{see further when clear}, thus, we propose the Curriculum Consistency Model (CCM). CCM incorporates three key designs, which are 1. A reliable metric Knowledge Discrepancy of the Curriculum (KDC) for measuring the difference between student and teacher over the interval $(t, u)$, 2. Dynamic adjustment of learning objectives based on the KDC, and 3. Multi-step iterative generation to ensure the quality of learning objectives.

\paragraph{Measuring the knowledge discrepancy.} We propose KDC based on PSNR to measure the knowledge discrepancy in Eq.~\ref{eq:KDC}. We have analyzed and shown the stability and generalizability of KDC across different datasets, different timesteps, and different training iterations in Section 3.

\paragraph{Dynamic adjustment of learning objectives.} To maintain the consistency of knowledge discrepancy across different timesteps and training iterations, we change the output of teacher model $\boldsymbol{x}_{\mathrm{target}}$ to $\boldsymbol{x}_{\mathrm{target}}^{\mathrm{KDC}}$. At each timestep, we cycle between estimating the knowledge discrepancy and modifying $u$ until the knowledge discrepancy exceeds a certain fixed value. At different values of $t$ and during various training iterations, we may obtain different values of $u$, showing the adaptive nature of CCM. Dynamic adjustment becomes effective at larger timesteps during the early stages of training, and extends across all timesteps in the later stages as the model progresses. Limited knowledge discrepancy results in a larger distillation step $l=u-t$ and allows the student to step further, avoiding cumulative errors from many small-step distillations and achieving improved image details, image-text alignment, and semantic structure.

\begin{figure}[H] 
\centering
\begin{minipage}[t]{0.36\textwidth} 
    \vspace{-5.2cm}
    \centering
    \includegraphics[width=\linewidth]{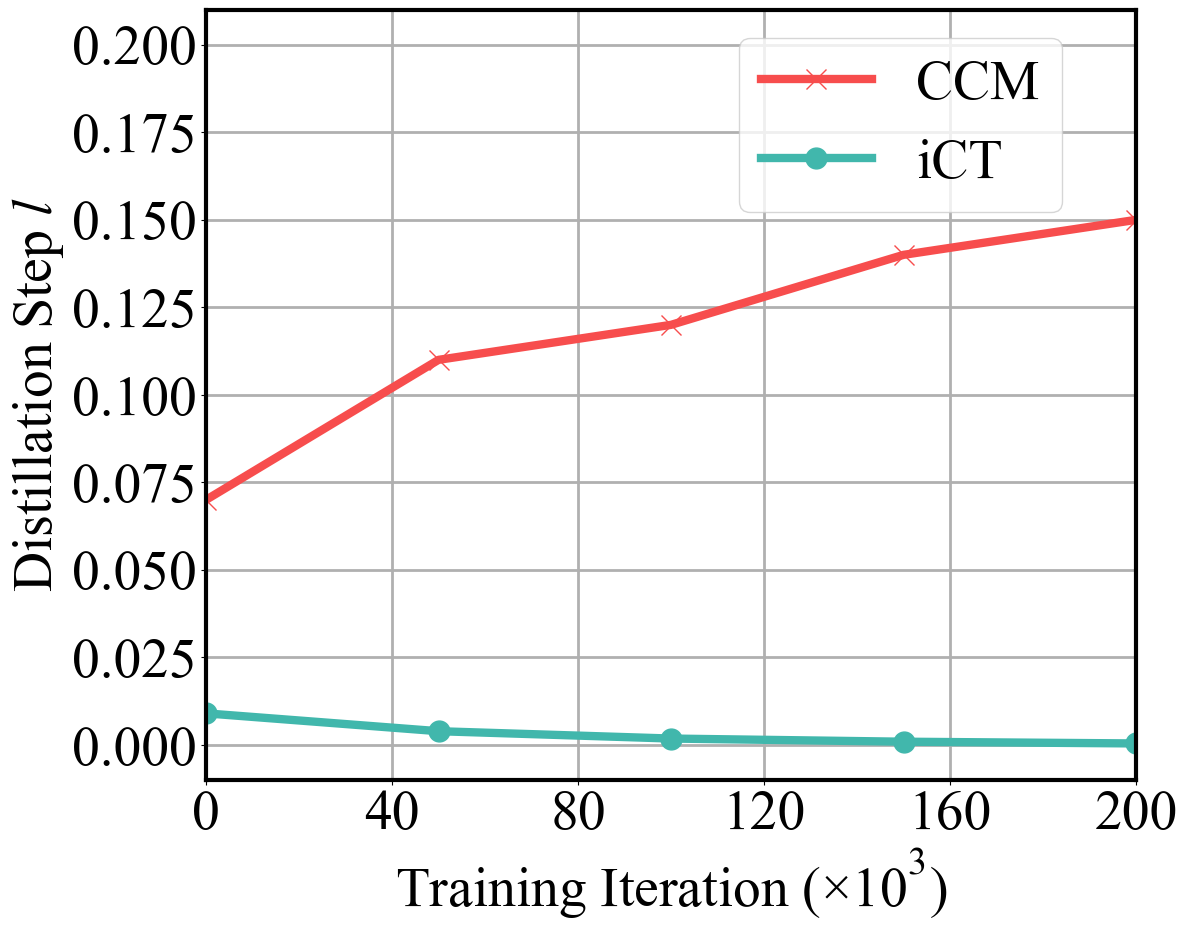}
    \caption{Distillation step vs. training iterations in CCM and iCT.}
    \label{fig:iteration}
\end{minipage}%
\hfill 
\begin{minipage}[b]{0.6\textwidth} 
    \centering
    \begin{algorithm}[H]
        \begin{algorithmic}[1]
            \STATE \textbf{Input:} noisy input \( \boldsymbol{x}_t \), timestep size \( s \), threshold \( T_{\mathrm{KDC}} \), teacher model \( \phi \), target model \( \boldsymbol{f}_{\theta^-} \), student model \( \boldsymbol{f}_{\theta} \)
            \STATE \textbf{Output:} KDC-Adjusted target \( \boldsymbol{x}_{\mathrm{target}}^{\mathrm{KDC}} \)
            
            \STATE Sample \( t \sim \mathcal{U}(0, 1) \)
            \STATE Calculate \( \boldsymbol{x}_{\mathrm{est}} = \boldsymbol{f}_{\theta}(\boldsymbol{x}_t, t, 1) \)
            
            \REPEAT
                \STATE Update \( u \gets \min(t + s, 1) \)
                \STATE Calculate \( \boldsymbol{x}_u = \mathrm{Solver}(\boldsymbol{x}_t, t, u; \phi) \)
                \STATE Compute \( \boldsymbol{x}_\mathrm{target}^{\mathrm{KDC}} = \boldsymbol{f}_{\theta^-}(\boldsymbol{x}_u,  u, 1) \)
                \STATE Compute \( \mathrm{KDC}_t^u = 100 - \mathrm{PSNR}(\boldsymbol{x}_{\mathrm{est}}, \boldsymbol{x}_{\mathrm{target}}^{\mathrm{KDC}}) \)
                \STATE Update \( t \gets u \), \( \boldsymbol{x}_t \gets \boldsymbol{x}_u \)
            \UNTIL \( T_{\mathrm{KDC}} < \mathrm{KDC}_t^u \) or \( u == 1 \)
        \end{algorithmic}
        \caption{KDC-Adjusted Target Computation}
        \label{alg:psnr_adjusted_target}
    \end{algorithm}
\end{minipage}
\end{figure}

\paragraph{Multi-step iterative generation.} Since the teacher model $\phi$ will remain the same in the training process, the pivotal issue for generating the learning objective at timestep $t$: \( \boldsymbol{x}_\mathrm{target} = \boldsymbol{f}_{\theta^-}(\mathrm{Solver}(\boldsymbol{x}_t,t,u;\phi),u,1) \) is to determine $\boldsymbol{x}_u = \mathrm{Solver}(\boldsymbol{x}_t,t,u;\phi)$. There are various methods to compute $\boldsymbol{x}_u$ and a straightforward approach is to estimate \( \boldsymbol{x}_u \) directly from \( \boldsymbol{x}_t \) through one-step iteration without regard for the magnitude of the distillation step $l=u-t$. However, CCM may select a \( u \) that is significantly greater than \( t \) to ensure a stable knowledge discrepancy, which could lead to the teacher model making inaccurate predictions due to a large timestep size $s$. Consequently, this may result in the student model learning targets that are vague or inaccurate. Therefore, we propose a multi-step iterative generation method where the teacher model will iterate one small timestep size $s$ forward each time until the estimated knowledge discrepancy meets the requirements, which are currently unknown. As shown in Figure~\ref{fig:iteration}, in contrast to iCT \cite{song2023improved}, CCM will increase the $l$ as training progresses.
Unlike the multi-step sampling in Scott \cite{liu2024scott}, where a large distillation step is subdivided and the relative positions of $u$ and $t$ remain fixed to reduce cumulative error, CCM determines $u$ by iterating forward from $t$. CCM allows the relative positions of $u$ and $t$ to vary dynamically across different timesteps and training iterations, ensuring the consistency of KDC.
For clarity, we have written the CCM algorithm's procedure in pseudocode and presented it in Algo\ref{alg:psnr_adjusted_target}.

\subsection{Unified Distillation Loss of CCM}

CCM focuses on addressing general issues in CM, thus making it applicable to a variety of common denoising-based generative models, including diffusion models and flow matching models. 
Suppose the ODE is defined on the time interval [0, 1) with 0 and 1 corresponding to noise and ground truth respectively, we can express the consistency distillation loss of CCM in a general form:

\begin{equation}
\label{eq:CCM}
\begin{split}
    \mathcal{L}_{\mathrm{CCM}}(\theta;\phi) := 
    \mathbb{E}_{t \in [0,1)}\mathbb{E}_{u \in (t, 1]}\mathbb{E}_{\boldsymbol{x}_{1}}\mathbb{E}_{\boldsymbol{x}_t|\boldsymbol{x}_{1}}[d(\boldsymbol{f}_{\theta}(\boldsymbol{x}_t,t,1), \boldsymbol{x}_{\mathrm{target}}^{\mathrm{KDC}}(u, 1))].
\end{split}
\end{equation}

where $t$ and $u$ are two timesteps of different noise intensities, $d(\cdot,\cdot)$ is a distance metric which can be L1, L2 or LPIPS. 
The difference between $\mathcal{L}_{\mathrm{CCM}}$ and standard consistency distillation loss is that the learning target $\boldsymbol{x}_{\mathrm{target}}^{\mathrm{KDC}}(u, 1)$ is obtained through a multi-step iteration according to Algo~\ref{alg:psnr_adjusted_target}.

\paragraph{CCM with diffusion models.} In diffusion models, it is customary to describe the denoising process using noise-to-signal ratio $\sigma \in [\epsilon, T]$, which can be transformed to timestep in Eq.~\ref{eq:CCM} through $t=\frac{1}{\sigma+1}$. The interval $[\epsilon, T]$ will be discretized firstly and standard consistency distillation loss can be calculated based on $\sigma_n$ and adjacent $\sigma_{n+1}$ as shown in Eq~\ref{eq:cd_loss}. CCM tends to calculate loss based on $\sigma_n$ and $\sigma_{n+m}$, where $m$ is the number of iteration steps according to Algo~\ref{alg:psnr_adjusted_target}.

\paragraph{CCM with flow matching models.} In flow matching models, a direct approach is to transform the noise-to-signal ratio $\sigma$ into discrete timesteps $t$ for consistency distillation, where $t$ becomes discrete within the range $0 < t = \frac{1}{\sigma+1} < 1$. We adopt an approach starting from vanilla flow matching, where $t$ is chosen uniformly within $[0, 1)$ \cite{lipman2023flow}. This approach leverages $t$ as a continuous variable, allowing consistency distillation to span a broader range of the ODE trajectory compared to discretized methods in diffusion models. Moreover, distillation at $t=0$ aligns with inference since generation begins from pure noise. Recent work in \cite{lu2024simplifying} also explores continuous-time consistency models. However, the selection of $u$ remains an open question. CCM offers a straightforward method to determine $u$ through adaptive iteration using a base timestep size $s$. In the following sections, we discuss the choice of $s$, $u$, and extra computational cost due to multi-step iterations.

\subsection{Adversarial Losses}
In generative modeling, student models derived from distillation often produce lower-quality samples compared to their teacher models, as they rely solely on distillation losses. To improve the student's performance and potentially surpass the teacher in quality, we incorporate adversarial training into our framework. Previous work, such as \cite{esser2021taming} and \cite{kim2023consistency}, has demonstrated that combining reconstruction and adversarial losses significantly enhances image generation quality.

Our Curriculum Consistency Model (CCM) framework integrates both KDC-adjusted distillation loss and adversarial losses into a unified training objective:

\begin{equation}
\begin{split}
    \mathcal{L}_{\mathrm{GAN}}(\theta, \eta) = & \mathbb{E}_{\boldsymbol{x}_1}(\log \boldsymbol{d}_{\eta}(\boldsymbol{x}_1) + 
    \mathbb{E}_{t \in [0,1)}\mathbb{E}_{\boldsymbol{x}_1}\mathbb{E}_{\boldsymbol{x}_t|\boldsymbol{x}_1}[\log(1 - \boldsymbol{d}_{\eta}(\boldsymbol{x}_{\mathrm{est}}(\boldsymbol{x}_t, t, 1)]
\end{split}
\end{equation}
\begin{equation}
    \min\limits_{\theta}\max\limits_{\eta} \mathcal{L}(\theta,\eta)=\mathcal{L}_{\mathrm{CCM}}(\theta;\phi) +\boldsymbol{\lambda}_{\mathrm{GAN}}\mathcal{L}_{\mathrm{GAN}}(\theta,\eta)
\end{equation}

where $\boldsymbol{d}_\eta$ represents the discriminator network and $\boldsymbol{\lambda}_{\mathrm{GAN}}$ is an adaptive weighting. Details are in \cite{kim2023consistency}.
\section{Experiments}

To verify the reliability and generalization of the method, our experiments cover classical datasets with different resolutions, and studies are carried out on diffusion models and flow matching models.

\subsection{Experimental Details}
\textbf{Datasets}. For low-resolution image generation, we train models on CIFAR-10 \cite{krizhevsky2009learning} and ImageNet 64x64 \cite{deng2009imagenet} datasets and evaluate them on the same datasets.
For high-resolution image generation, we train LoRA weights \cite{hu2022lora} on the CC3M \cite{changpinyo2021conceptual} dataset and evaluate on COCO-2017 \cite{lin2014microsoft} with our chosen 5K split.

\textbf{Models}. We verify the image generation based on both flow matching and diffusion models, including Optimal Transport Conditional Flow Matching (OT-CFM)~\cite{tong2023improving}, Stable Diffusion 3~\cite{esser2024scaling}, and Stable Diffusion XL~\cite{podell2024sdxl}. 
Our code implementation is based on torchcfm and phased consistency model \cite{pcm2024wang}.

\textbf{Evaluation Metrics}. We report the FID \cite{heusel2017gans} and CLIP Score \cite{radford2021learning} of the generated images and the validation 5K-sample splits. We also comprehensively evaluate the compositionality of CCM on T2I-CompBench \cite{huang2023t2i}.

Our experimental parameters are shown in the Appendix.

\subsection{Experimental Results and Analysis}

\begin{table}[h]
  \centering

  \renewcommand{\arraystretch}{1.2}

  \begin{minipage}[t]{0.48\linewidth}
    \centering
    \begin{table}[H]
      \centering
      \captionsetup{font=small}
      \caption{Performance comparisons on CIFAR-10}
      \resizebox{\columnwidth}{!}{%
      \begin{tabular}{l|l r l}
        \hline
        \textbf{Model Type} & \textbf{Method} & \textbf{NFE} ($\downarrow$) & \textbf{FID} ($\downarrow$) \\
        \hline
        \multirow{1}{*}{GAN} & StyleGAN-XL(\cite{sauer2022stylegan})             & 1     &   1.85    \\
        \hline
        \multirow{10}{*}{Diffusion Models} & DDPM(\cite{ho2020denoising})                      & 1000  &   3.17    \\
        & DDIM(\cite{songdenoising})                        & 100   &   4.16    \\
        & Score SDE(\cite{songscore})                       & 2000  &   2.20    \\
        & EDM(\cite{karras2022elucidating})                 & 35    &   2.01    \\
        & 2-Rectified Flow(\cite{liu2023flow})              & 1     &   4.85    \\
        & ECM(\cite{geng2024consistency}) & 1 & 3.60 \\
        & CD(\cite{song2023consistency})                    & 1     &   3.55    \\
        & iCT(\cite{song2023improved}) & 1 & 2.83 \\
        & CD + GAN(\cite{lu2023cm})                         & 1     &   2.65    \\
        & CTM(\cite{kim2023consistency})                    & 1     &   1.98    \\
        \hline
        \multirow{3}{*}{Flow Matching Models} & OT-CFM(\cite{tong2023improving})                    & 100     &   4.49    \\
        & PCM(\cite{pcm2024wang})                           & 8     &   1.94    \\
        & \textbf{CCM (ours)}                               & 1     &   \textbf{1.64}    \\
        \hline
      \end{tabular}
      }
      \label{tab:cifar10}
    \end{table}
  \end{minipage}%
  \hfill
  \begin{minipage}[t]{0.48\linewidth}
    \centering
    \begin{table}[H]
      \centering
      \captionsetup{font=small}
      \caption{Performance comparisons on ImageNet 64×64}
      \resizebox{\columnwidth}{!}{%
      \begin{tabular}{l|lll}
      \hline
        \textbf{Model Type} & \textbf{Method} & \textbf{NFE} ($\downarrow$) & \textbf{FID} ($\downarrow$) \\
        \hline
        \multirow{5}{*}{Diffusion Models} & EDM(\cite{karras2022elucidating})                     & 79    &   2.44            \\
        & CD(\cite{song2023consistency})                        & 1     &   6.20            \\
        & ECM(\cite{geng2024consistency}) & 1 & 4.05 \\
        & iCT(\cite{song2023improved}) & 1 & 4.02 \\
        & CTM(\cite{kim2023consistency})                        & 1     &   \textbf{1.92}   \\
        \hline
        \multirow{2}{*}{Flow Matching Models} & OT-CFM(retrained)        & 100   &   5.36            \\
        & \textbf{CCM (ours)}                                           & 1     &   \textbf{2.18}            \\
        \hline
      \end{tabular}
      }
      \label{tab:imgenet64}
    \end{table}
  \end{minipage}

  \vspace{-1cm} 
  \begin{minipage}[b]{0.48\linewidth}
    \centering
    \begin{table}[H]
      \centering
      \captionsetup{font=small}
      \caption{Performance comparisons on CoCo2017-5K}
      \resizebox{\columnwidth}{!}{%
      \begin{tabular}{l|lll}
        \hline
        \textbf{Base Model} & \textbf{Method} & \textbf{CLIP Score} ($\uparrow$) & \textbf{FID} ($\downarrow$) \\
        \hline
        \multirow{4}{*}{SD3} & Original                  & 28.09     & 99.61      \\
        & LCM(\cite{luo2023latent})                   & 32.32     & 35.62      \\
        & PCM(\cite{pcm2024wang})                   & 32.34     & 33.22      \\
        & \textbf{CCM(ours)}                      & \textbf{32.42}     & \textbf{32.54}      \\
        \hline
        \multirow{4}{*}{SDXL} & Original                   & 30.41     & 70.28      \\
        & Hyper-SD(\cite{ren2024hyper})           & 32.10 &  30.38 \\ 
        & PCM(\cite{pcm2024wang})                  & 32.47     & 29.89      \\
        & \textbf{CCM(ours)}                                  & \textbf{32.60}     & \textbf{28.90}      \\
        \hline
      \end{tabular}
      }
      \label{tab:coco}
    \end{table}
  \end{minipage}%
  \hfill
  \begin{minipage}[b]{0.48\linewidth}
    \centering
    \includegraphics[width=0.8\linewidth]{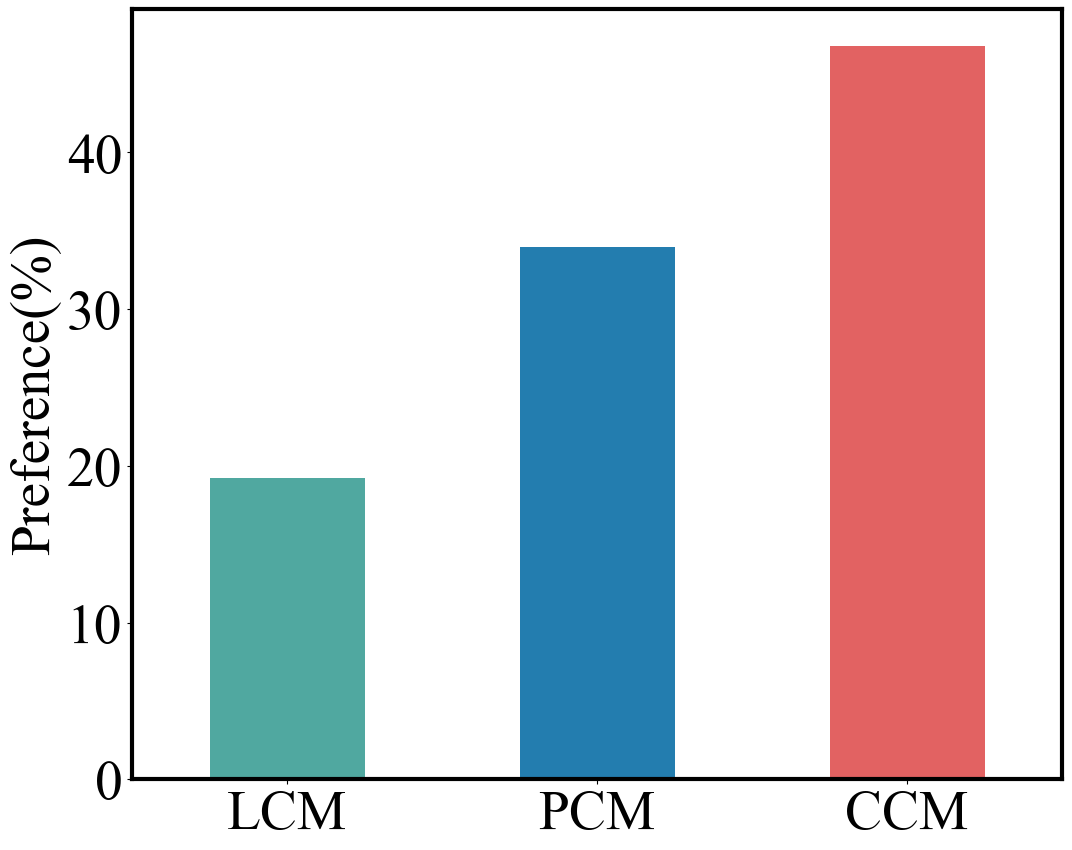} 
    \captionsetup{type=figure}
    \caption{User study. Subjects were shown generated images and asked for preference.}
    \label{fig:user}
  \end{minipage}
\end{table}

\begin{table*}[h]
    \centering
    \caption{Quantitative Results on T2I-CompBench \cite{huang2023t2i}. CCM provides consistent improvements in all categories for both SD3 and SDXL. \colorbox{cyan!20}{Blue} means the reference results from the original models (28 steps for SD3 and 40 steps for SDXL). Other models use 4 inference steps.}
    \label{table:bench}
    \resizebox{\linewidth}{!}{%
        \begin{tabular}{@{}llccccccc@{}}
        \toprule
        \multirow{2}{*}{\textbf{Base Model}} & \multirow{2}{*}{\textbf{Method}} & \multicolumn{3}{c}{\textbf{Attribute Binding}} & \multicolumn{2}{c}{\textbf{Object Relationship}} & \multirow{2}{*}{\textbf{Complex} ($\uparrow$)} \\ 
        \cmidrule(lr){3-5} \cmidrule(lr){6-7}
        & & \textbf{Color} ($\uparrow$) & \textbf{Shape} ($\uparrow$) & \textbf{Texture} ($\uparrow$) & \textbf{Spatial} ($\uparrow$) & \textbf{Non-Spatial} ($\uparrow$) &  \\ 
        \midrule
        \rowcolor{cyan!20} \multirow{5}{*}{\cellcolor{white} SD3}
        & Original & 0.813 & 0.590 & 0.759 & 0.343 & 0.311 & 0.479 \\
        & LCM (\cite{luo2023latent}) & 0.705 &	0.482 &	0.587 &	0.225 & 0.309 & 0.346 \\
        & PCM (\cite{pcm2024wang}) & 0.702 &	0.480 &	0.599 &	0.212 &	0.305 &	0.346 \\
        & \textbf{CCM(ours)} & \textbf{0.733} &	\textbf{0.493} &	\textbf{0.633} &	\textbf{0.245} &	\textbf{0.310} &	\textbf{0.358} \\
        \midrule
        \rowcolor{cyan!20} \multirow{5}{*}{\cellcolor{white} SDXL}
        & Original & 0.587 & 0.468 & 0.529 & 0.213 & 0.311 & 0.323\\
        & LCM (\cite{luo2023latent}) & 0.604 & 0.407 & 0.497 & 0.172 & 0.310 & 0.337 \\
        & PCM (\cite{pcm2024wang}) & 0.606 & 0.420 & 0.497 & 0.202 & 0.311 & 0.332 \\
        & Lightning (\cite{lin2024sdxl}) & 0.581 & \textbf{0.437} & 0.499 & \textbf{0.221} & 0.311 & 0.325  \\
        & \textbf{CCM(ours)} & \textbf{0.614} & 0.427 & \textbf{0.511} & 0.207 & \textbf{0.312} & \textbf{0.338}  \\
        \bottomrule
        \end{tabular}
    }
\end{table*}

\begin{figure}[h]
\centering
\scriptsize  
\begin{tabular}{m{0.1cm}<{\centering} m{2.2cm}<{\centering}  m{2.2cm}<{\centering} m{2.2cm}<{\centering}}
        & OT-CFM \quad \quad \quad (NFE=1) & OT-CFM \quad \quad \quad (NFE=100) & CCM \quad \quad \quad (NFE=1) \\ 
\rotatebox{90}{CIFAR-10}
        & \includegraphics[width=\linewidth]{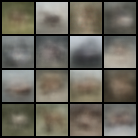} 
        & \includegraphics[width=\linewidth]{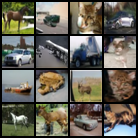} 
        & \includegraphics[width=\linewidth]{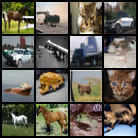} \\
\rotatebox{90}{ImageNet 64x64}
        & \includegraphics[width=\linewidth]{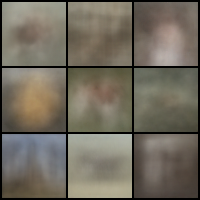} 
        & \includegraphics[width=\linewidth]{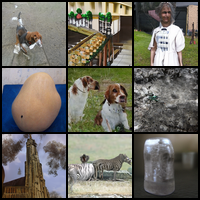} 
        & \includegraphics[width=\linewidth]{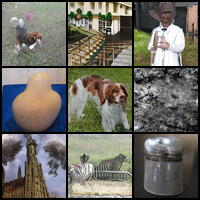} \\ 
\end{tabular}
\caption{Samples generated by OT-CFM and CCM on CIFAR-10 and ImageNet 64x64.}
\label{fig:lowres}
\end{figure}

Based on the experimental results provided in Table \ref{tab:cifar10}-\ref{tab:coco}, we conduct a performance analysis of the Curriculum Consistency Model (CCM) compared to existing approaches. 
On the CIFAR-10 dataset, CCM achieves an impressive unconditional FID of 1.64 with only one function evaluation (NFE=1), outperforming other methods.
CCM not only surpasses these methods in sampling efficiency but also achieves superior image quality.  
On the ImageNet 64×64 dataset, CCM also performed strongly: CCM’s FID (NFE=1) reaches 2.18 on conditional generation, which is also competitive with the mainstream generated models. 
Although the performance of a student model heavily depends on its teacher, CCM (2.18) demonstrates a more substantial improvement over its teacher model, OT-CFM (5.36), than CTM (1.92) does over its teacher model, EDM (2.44).
The samples generated by CCM (NFE=1) trained on CIFAR-10 and ImageNet 64x64 are shown in Figure \ref{fig:lowres}. CCM shows excellent acceleration that the images generated by CCM in one step are comparable in quality to those generated by OT-CFM in 100 steps, and at least 50x faster in inference. Additional images are provided in the appendix for further reference. The training cost of CCM will be discussed in ablation studies.

\begin{figure}[h]
    \centering
    \begin{minipage}[t]{0.48\linewidth}
        \centering
        \scriptsize  
        \setlength{\tabcolsep}{1pt}
        \resizebox{\linewidth}{!}{%
        \begin{tabular}{m{2cm}<{\centering}  m{2cm}<{\centering} m{2cm}<{\centering} m{2cm}<{\centering}}
                A \colorbox{cyan!20}{\textbf{coffee}} mug floating in the sky. 
                & An overhead view of a pickup \colorbox{cyan!20}{\textbf{truck}} with boxes in its flatbed.
                & A white flag with a red \colorbox{cyan!20}{\textbf{circle}} next to a solid blue flag.
                & A ceiling fan with \colorbox{cyan!20}{\textbf{five}} brown blades.\\ 
                \includegraphics[width=\linewidth]{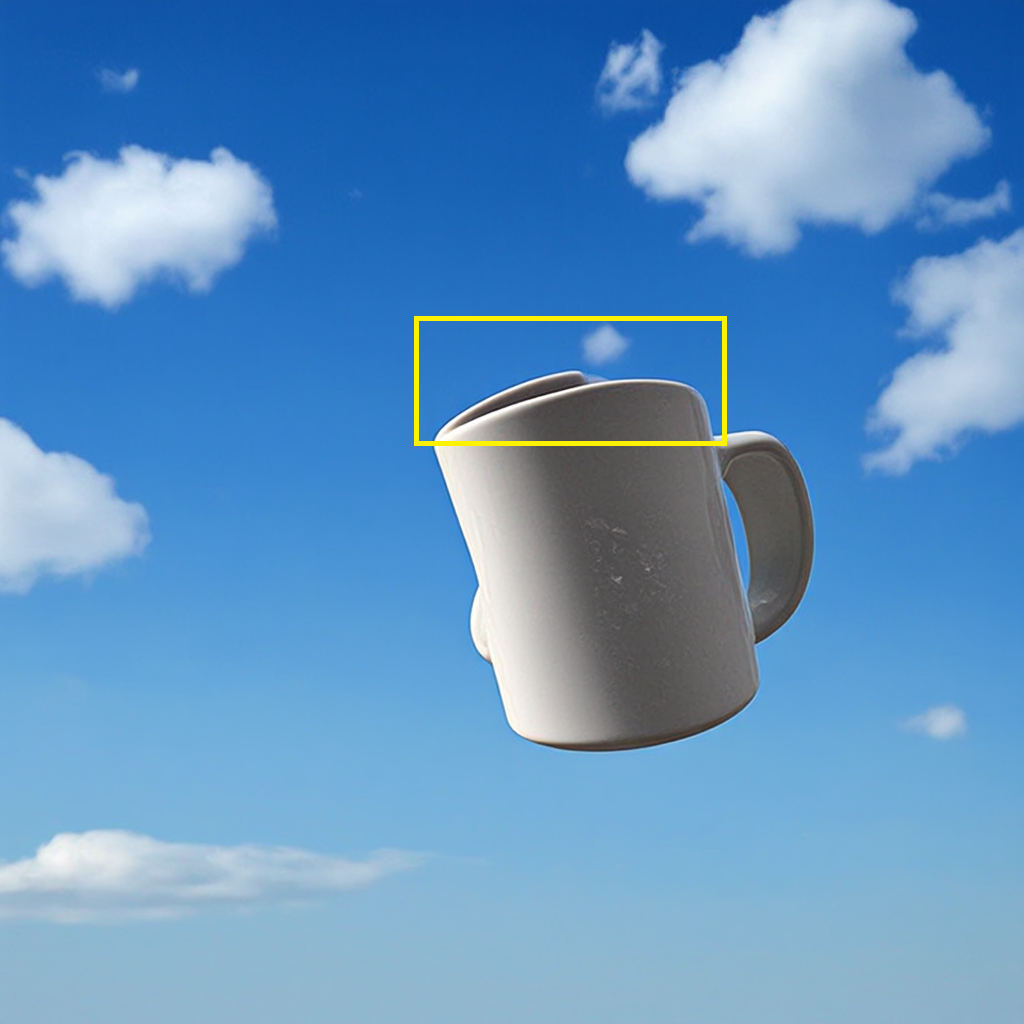} 
                & \includegraphics[width=\linewidth]{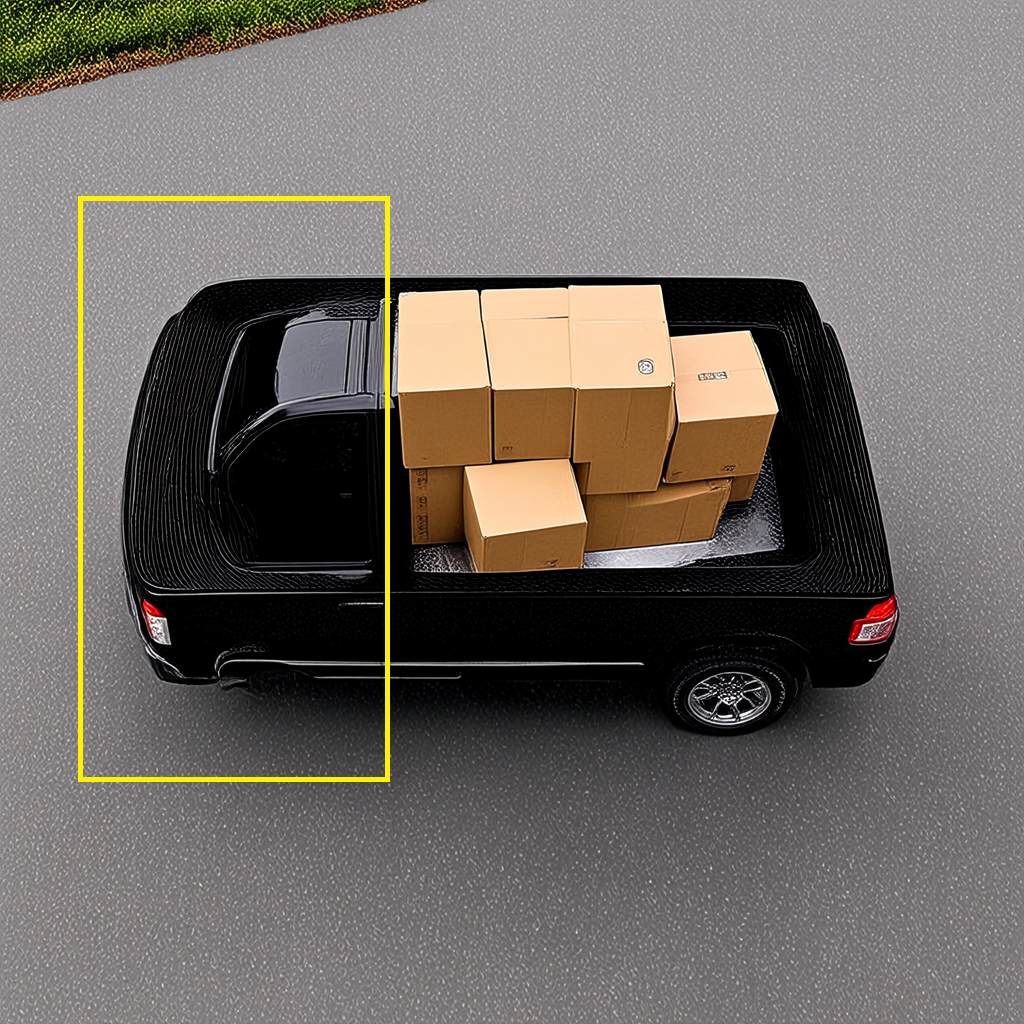} 
                & \includegraphics[width=\linewidth]{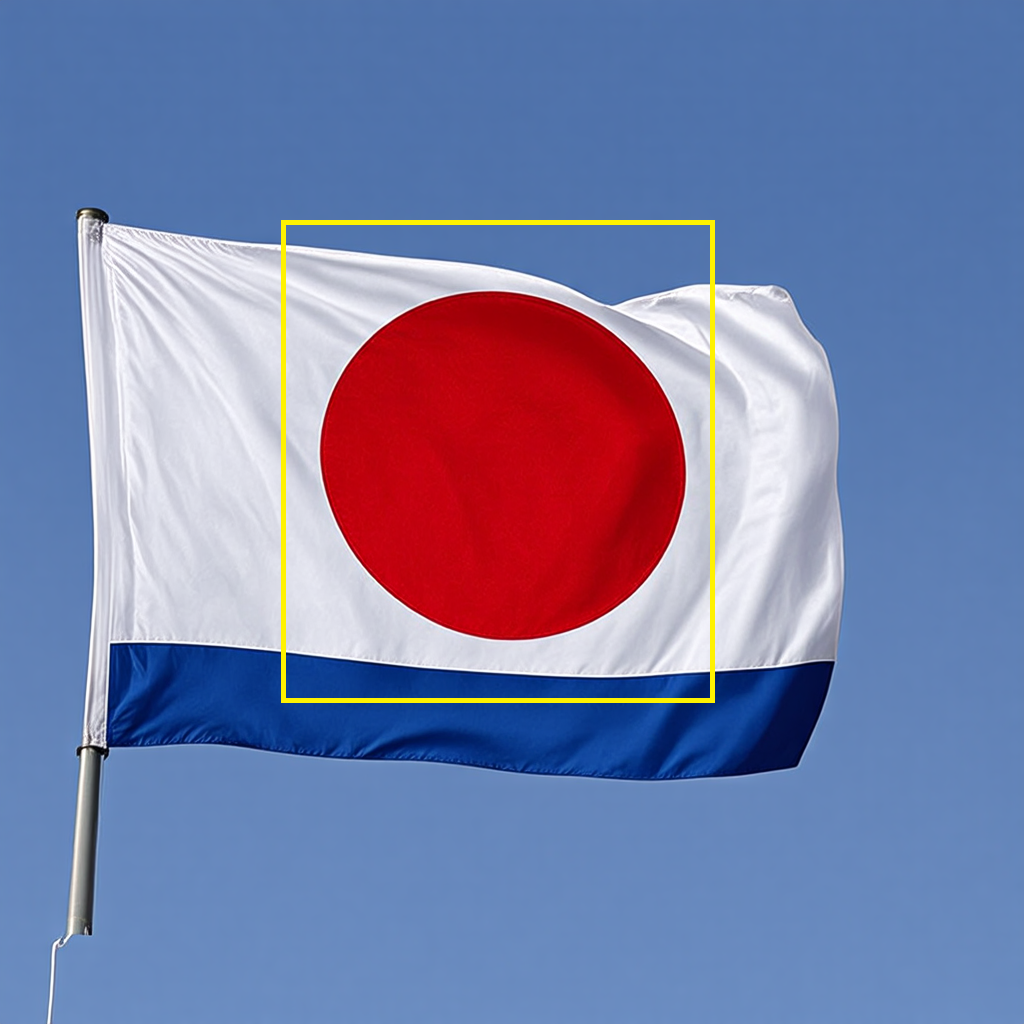} 
                & \includegraphics[width=\linewidth]{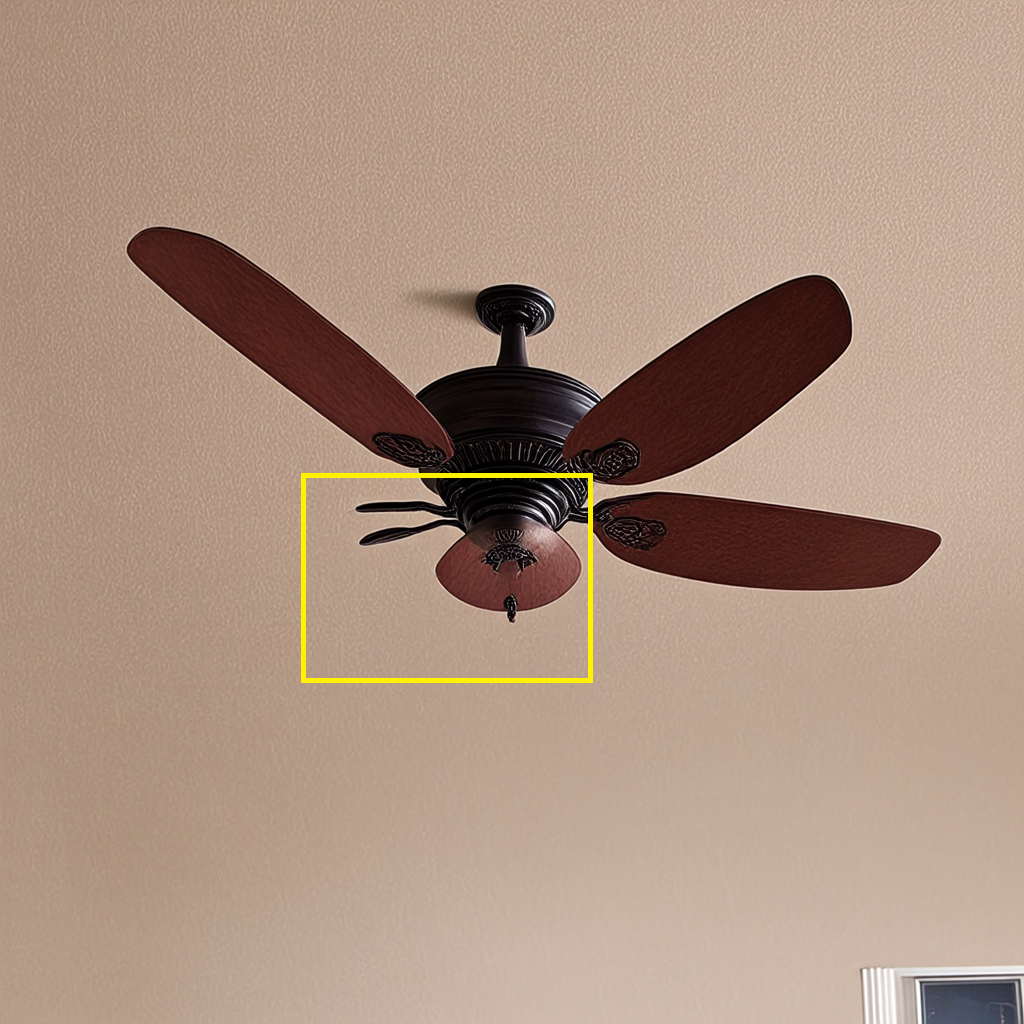} \\
                
                \includegraphics[width=\linewidth]{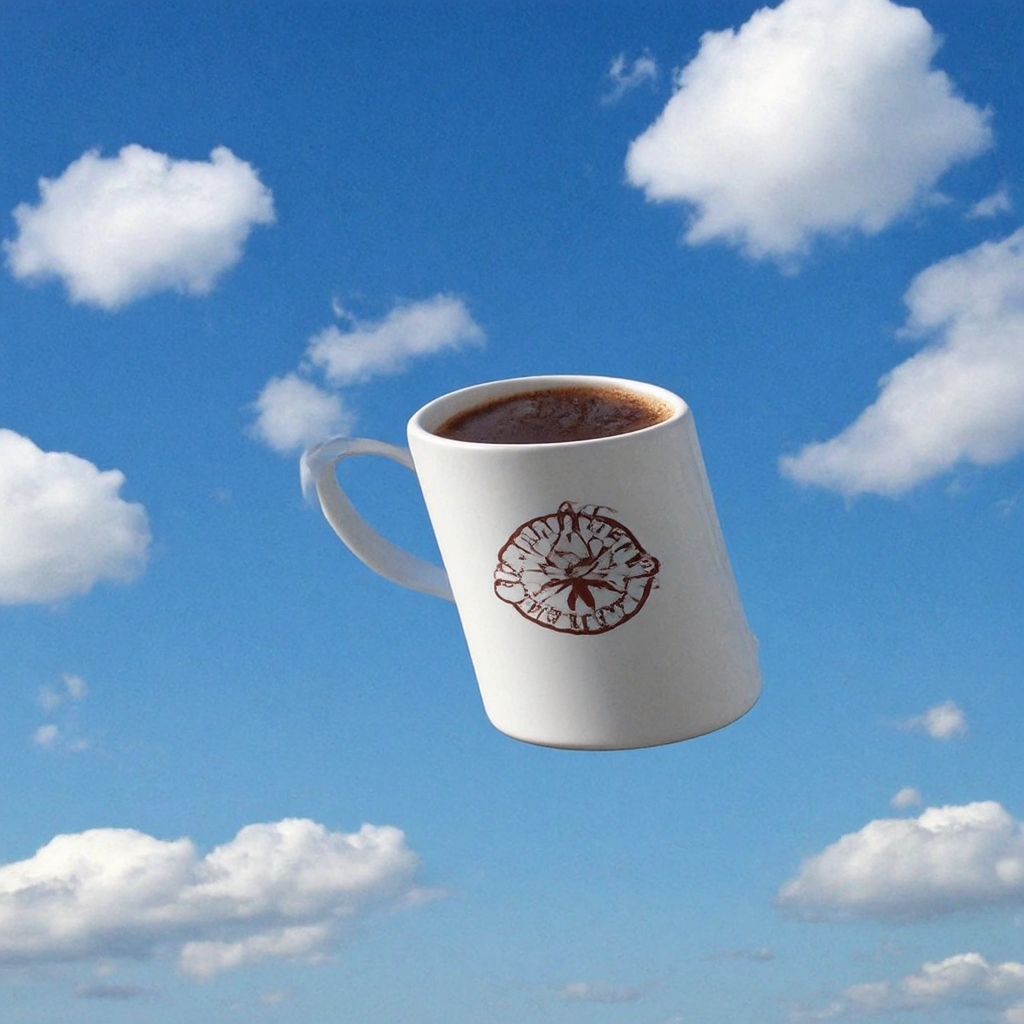} 
                & \includegraphics[width=\linewidth]{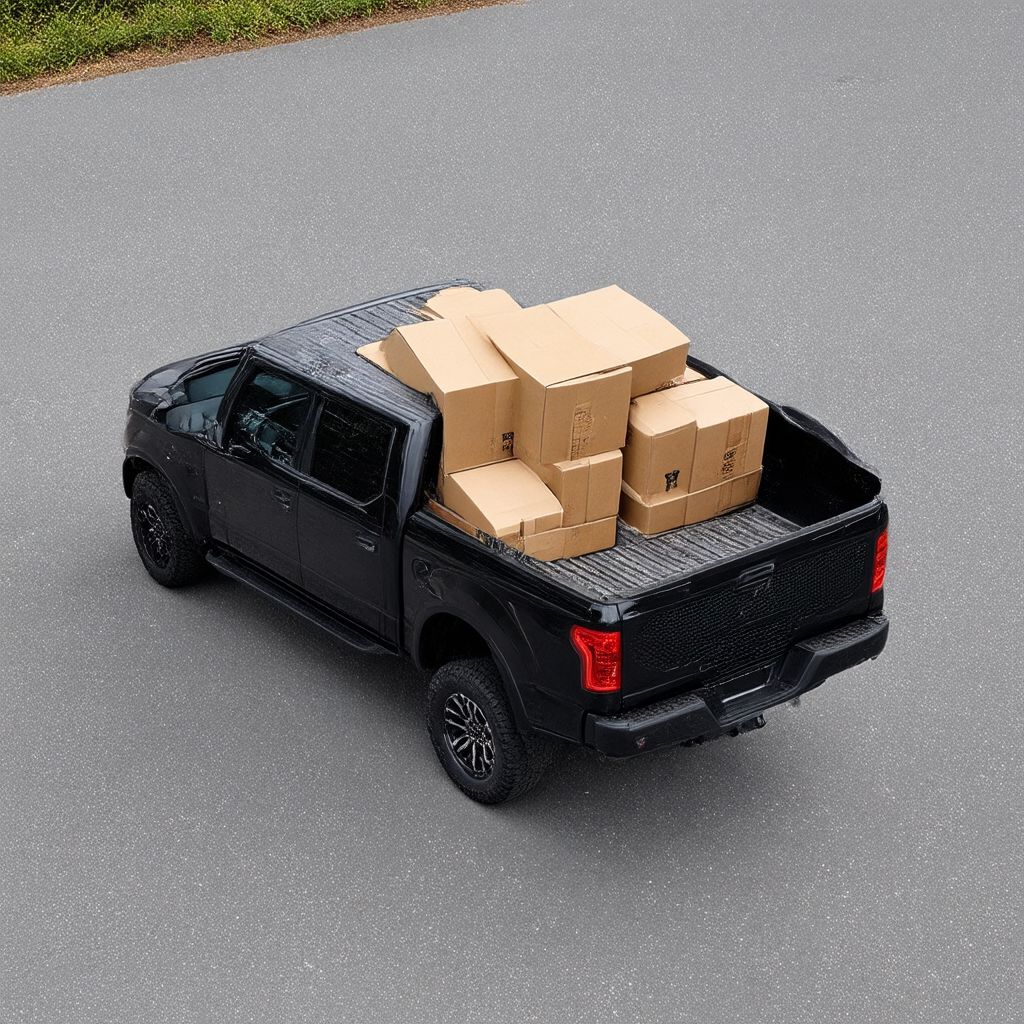} 
                & \includegraphics[width=\linewidth]{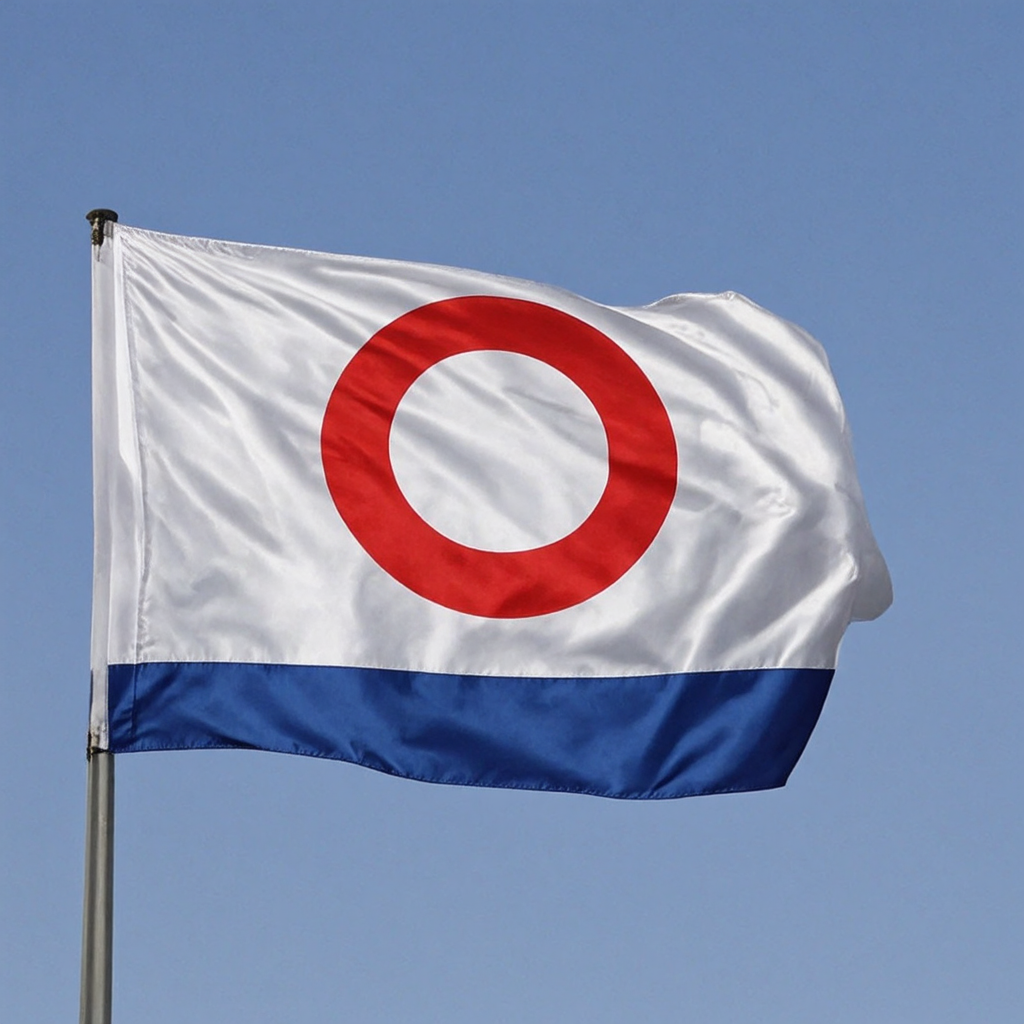} 
                & \includegraphics[width=\linewidth]{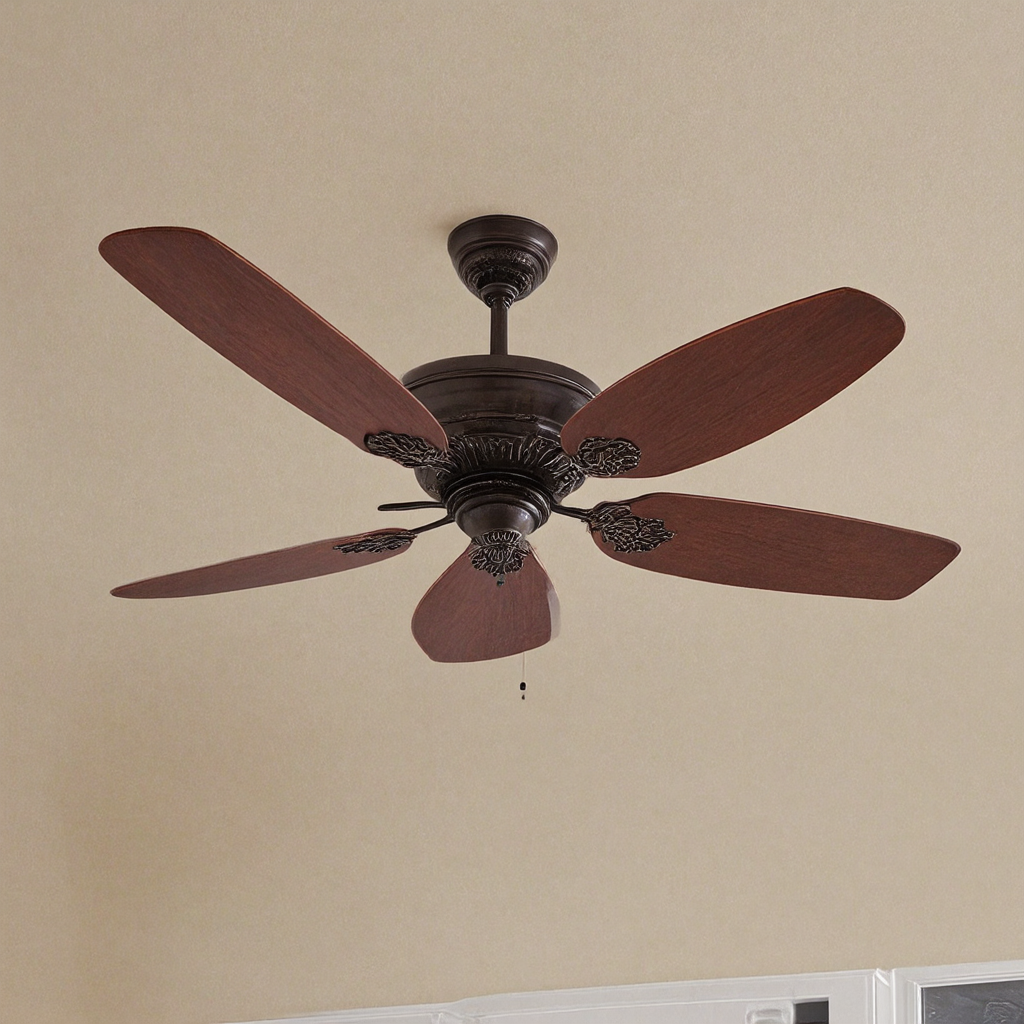} \\
        \end{tabular}
        }
        \caption{Semantic comparison of images generated by LCM (up) and CCM (down). CCM shows better image-text alignment and generates images that better fit the text.}
        \label{fig:sd_semantic}
    \end{minipage}
    \hfill
    \begin{minipage}[t]{0.48\linewidth}
        \centering
        \scriptsize  
        \setlength{\tabcolsep}{1pt}
        \resizebox{\linewidth}{!}{%
        \begin{tabular}{m{2cm}<{\centering}  m{2cm}<{\centering} m{2cm}<{\centering} m{2cm}<{\centering}}
                A glass of orange juice to the right of a plate with buttered toast on it.
                & A drawing of a stork playing a violin.
                & A grand piano next to the net of a tennis court.
                & A giraffe points its head towards the sky.\\ 
                \includegraphics[width=\linewidth]{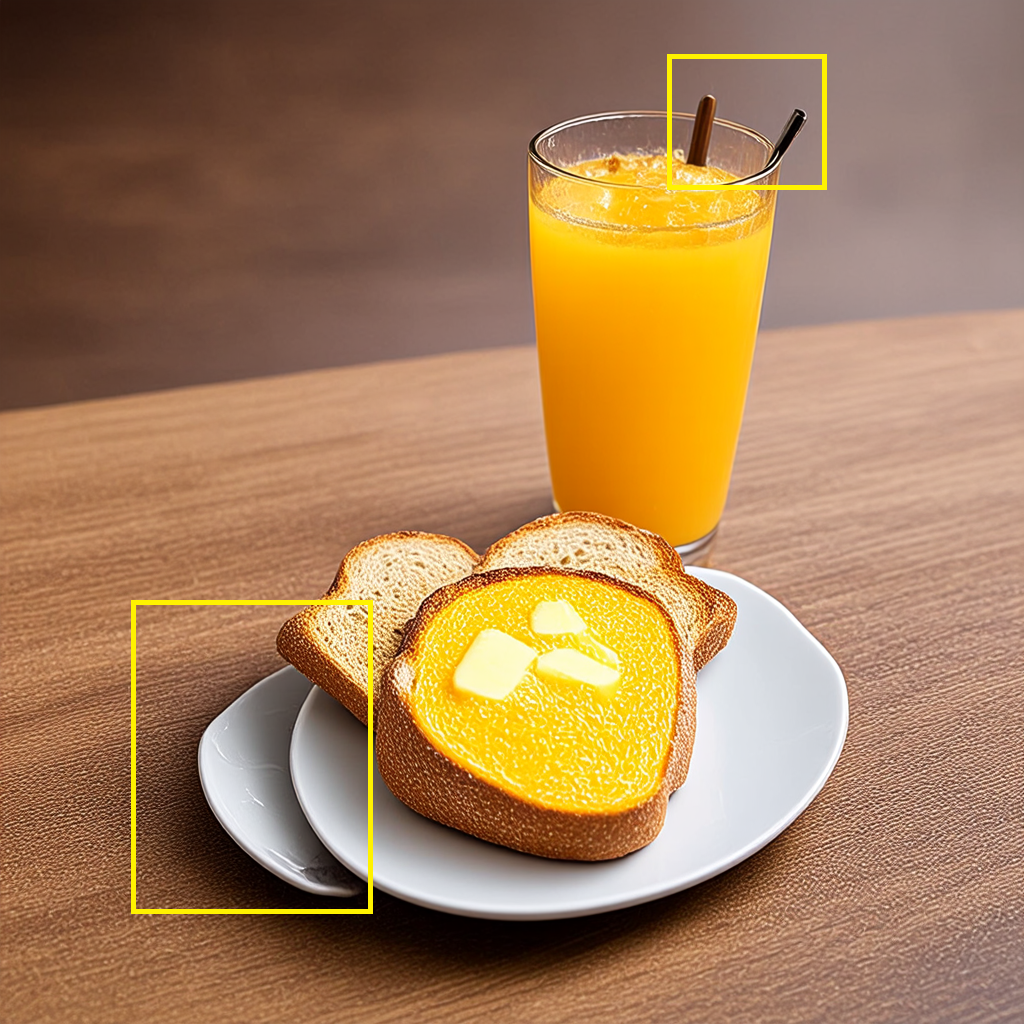} 
                & \includegraphics[width=\linewidth]{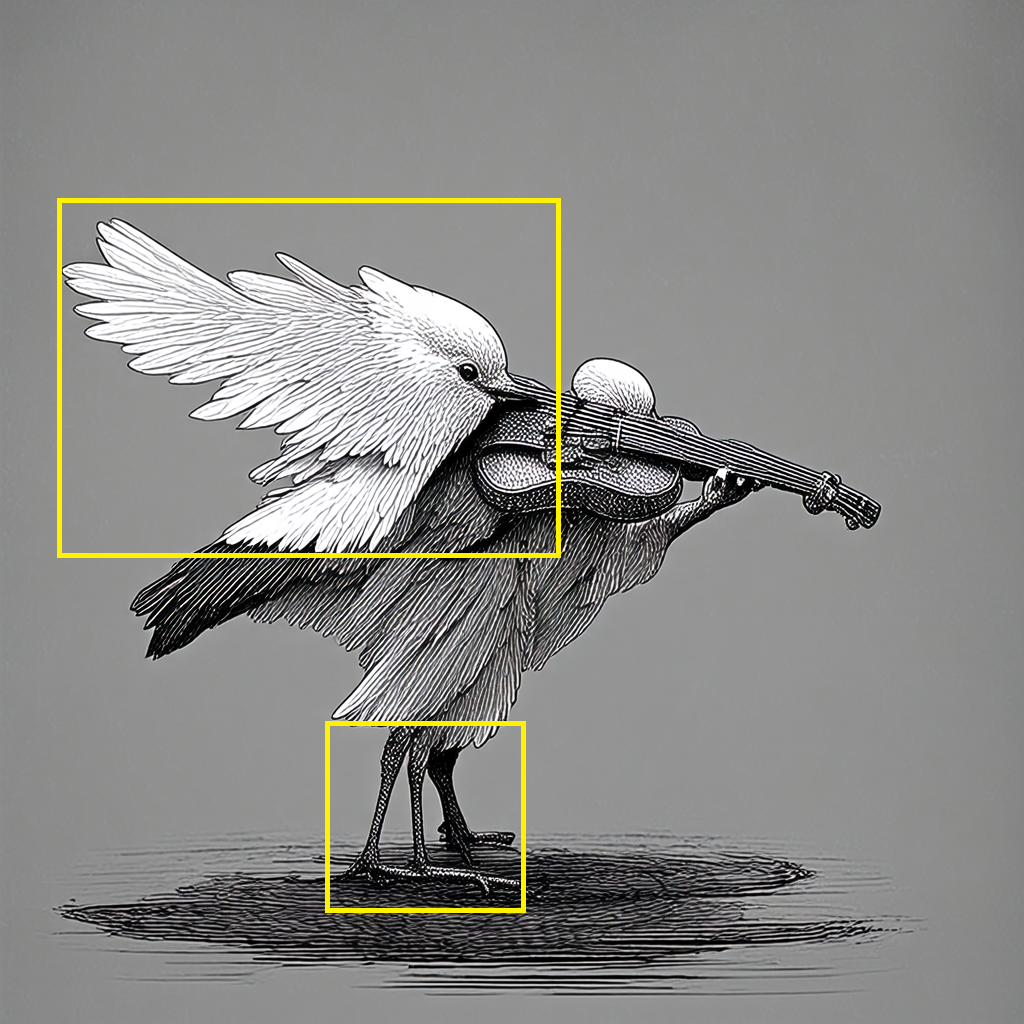} 
                & \includegraphics[width=\linewidth]{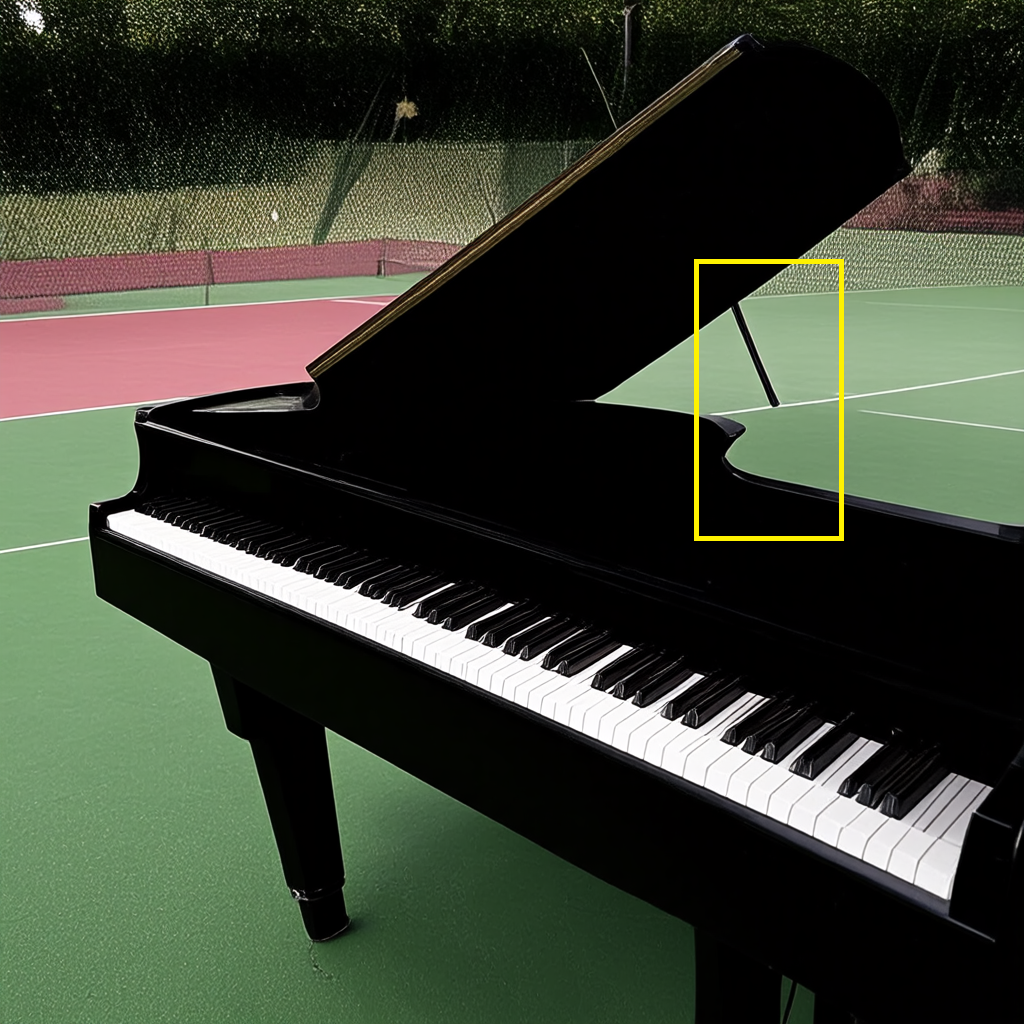} 
                & \includegraphics[width=\linewidth]{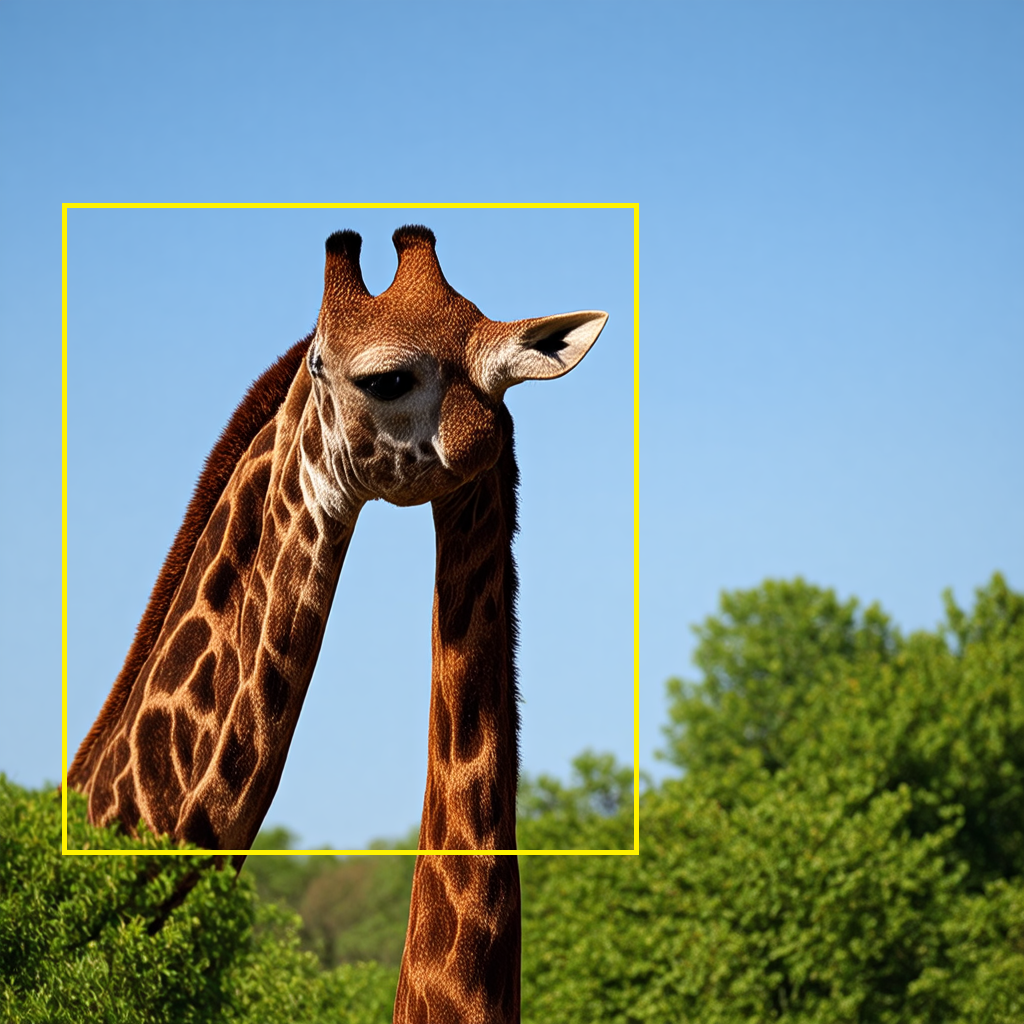} \\
                
                \includegraphics[width=\linewidth]{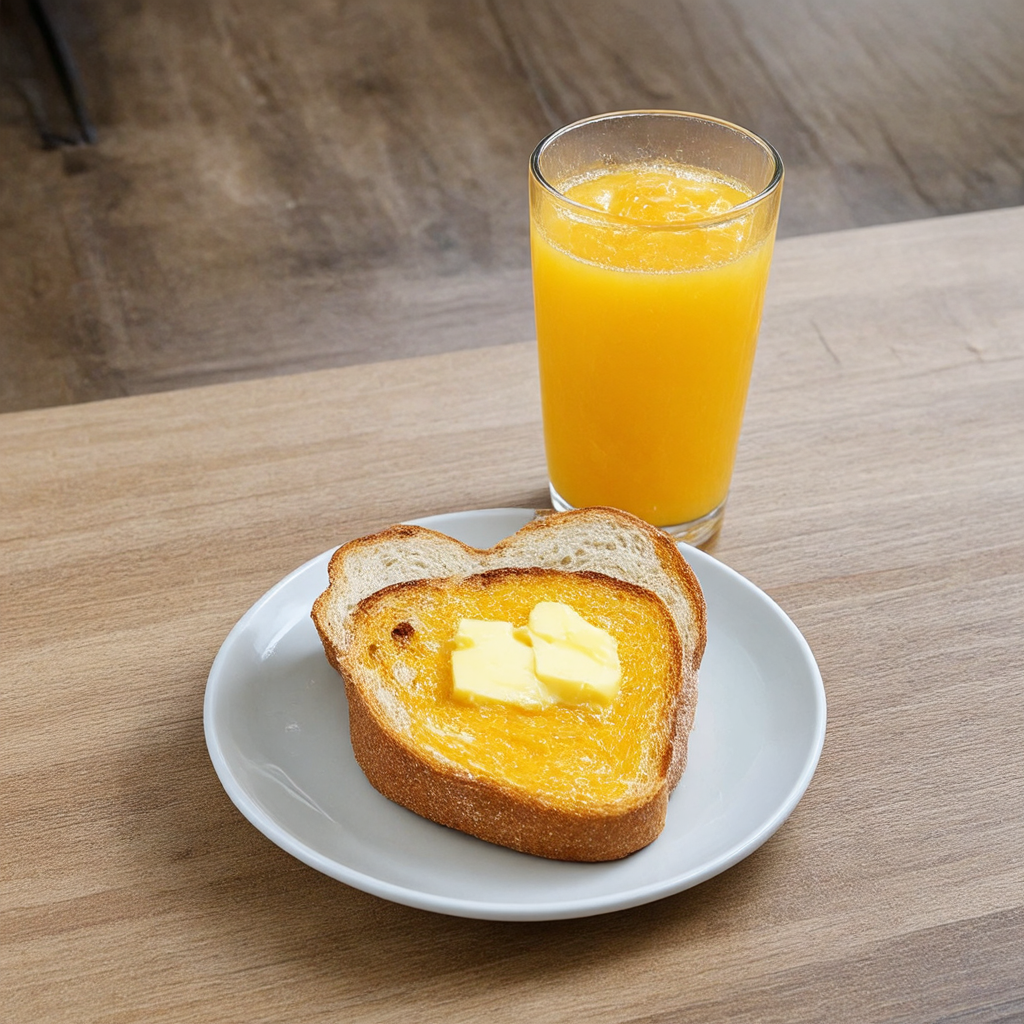} 
                & \includegraphics[width=\linewidth]{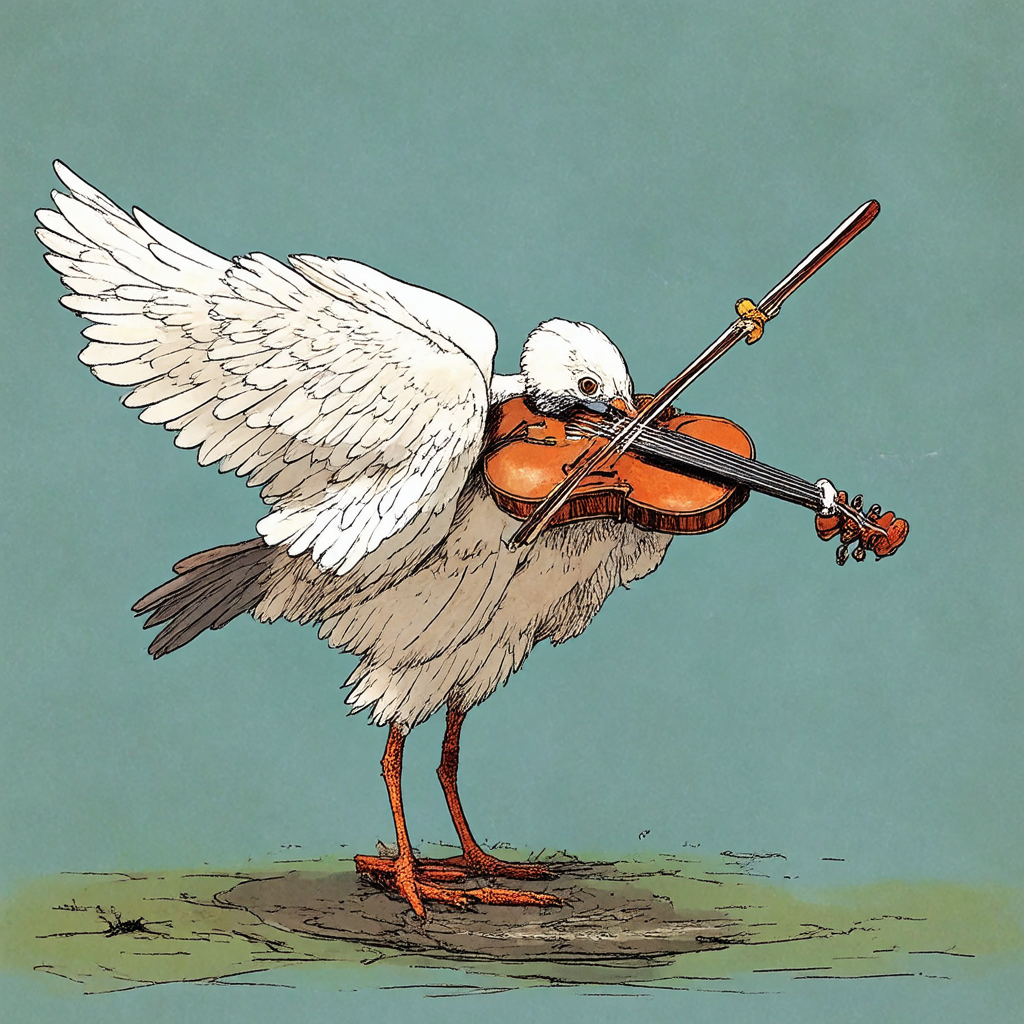} 
                & \includegraphics[width=\linewidth]{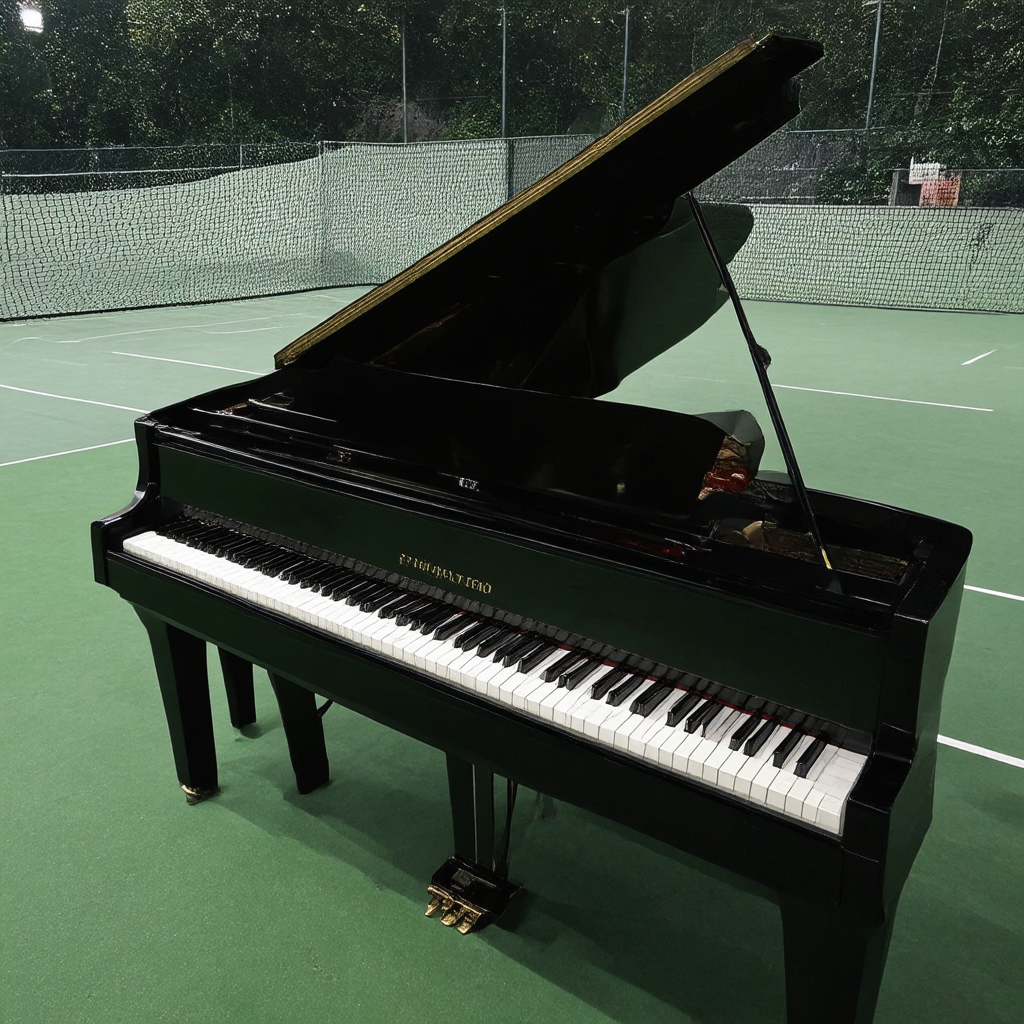} 
                & \includegraphics[width=\linewidth]{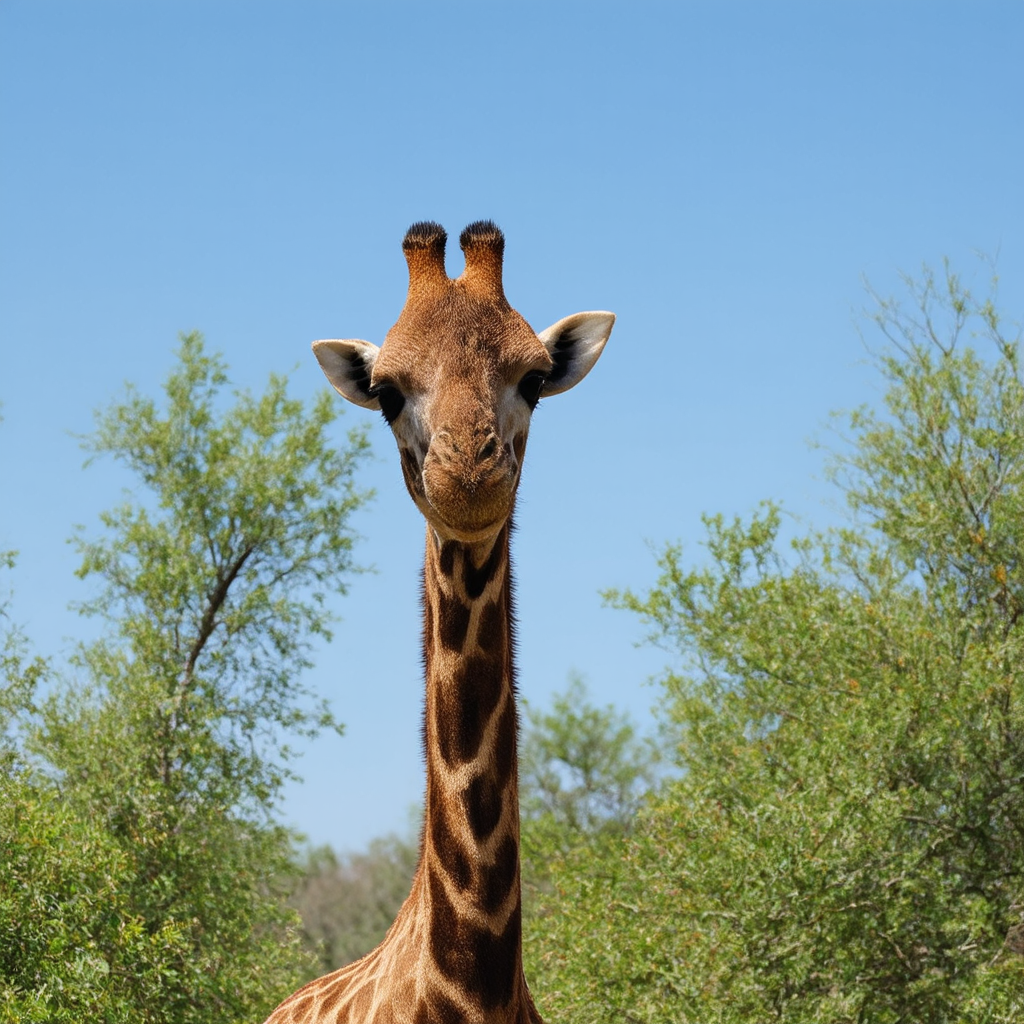} \\
        \end{tabular}
        }
        \caption{Structure comparison of images generated by LCM (up) and CCM (down). Both models correctly understand the text, but structures generated by CCM are more reasonable.}
        \label{fig:sd_structure}
    \end{minipage}
\end{figure}

When scaled to large-scale methods and high resolution, CCM can still maintain advantages. According to Table~\ref{tab:coco}, CCM has achieved lower FID and higher CLIP scores on both diffusion models and flow matching models. 
On T2I-Compbench \cite{huang2023t2i}, CCM-4Step outperforms both LCM and PCM across all six metrics, achieving results comparable to SD3-28Step. Additionally, CCM based on SDXL performs well in color, texture, non-spatial, and complex attributes. 
We compare the samples generated by different methods and find that CCM performs better image-text alignment (Figure. \ref{fig:sd_semantic}) and semantic structure (Figure. \ref{fig:sd_structure}). 
Further, we conduct a user study and Figure~\ref{fig:user} affirms the good performance of CCM.
The results demonstrate the strong generalization capabilities of CCM.

\subsection{Ablation Studies}

\begin{wraptable}{r}{0.4\textwidth} 
    \centering
    \vspace{-15pt}
    \caption{Comparison between static and dynamic strategies. For CCM, $T_{\mathrm{KDC}}=60$. I-CCM adopts the opposite strategy of CCM. $l = \sum_{i=1}^{n} s_i$.}
    \label{tab:static_dynamic}
    \resizebox{\linewidth}{!}{%
        \begin{tabular}{l|cccl}
            \hline 
            \textbf{Strategy} & $l$ & $n$ & $s$ & \textbf{FID} ($\downarrow$) \\
            \hline 
            \multirow{5}{*}{Static}  
            & $0.01$ & 1 & 0.01 & 14.06 \\
            & $0.03$ & 1 & 0.03 &  11.38 \\
            & $0.1$  & 1 & 0.1 & 16.2 \\
            & $0.06$ & 2 & 0.03 & 10.15 \\
            & $0.09$ & 3 & 0.03 & 9.89 \\ 
            \hline
            \multirow{4}{*}{Dynamic}
            & $0.1t$ & 1 & 0.1t & 27.19 \\
            & I-CCM & - & 0.03 & 12.66 \\
            & $1-t$ & 1 & 1-t & 10.67 \\
            & \textbf{CCM} & - & 0.03 & \textbf{9.32} \\
            \hline  
        \end{tabular}
    }
    \vspace{-20pt}
\end{wraptable}
We perform thorough ablation studies to evaluate the impact of different modules in the method. All ablation experiments are based on CIFAR-10 without adversarial losses.

\paragraph{Static vs. Dynamic.}
We first compared different target selection strategies to study the effect of the distillation step $l=u-t$, numbers of iterative steps $n$, and timestep sizes $s$. The three key variables have the following relation $l = \sum_{i=1}^{n} s_i$. Strategies fall into two categories: static strategies that $l, s, n$ are fixed and dynamic strategies that at least one variable in $l,s,n$ varies with $t$. 
From Table \ref{tab:static_dynamic}, we can observe that CCM surpasses all other strategies. 
Moreover, when $n$ increases from 1 to 3 with fixed $s=0.03$, the model's performance improves. Similarly, increasing the distillation step $l=u-t$ also exhibits a similar phenomenon, but a larger value of $l$ with fewer iterative steps $n$ can be detrimental($l=0.1, n=1$).
Furthermore, we experimented with varying the timestep size $s$ in accordance with the changes in \( t \). Increasing $l$ proportionally as \( t \) increases is not a good choice since it is almost impossible to learn when both $t$ and timestep size $s$ are very small, which also reminds us to balance knowledge discrepancy and model ability. A special case of the opposite is learning ground truth directly, i.e., $l=s=1-t$, which also lags behind CCM. Last, I-CCM, which uses an opposite strategy to CCM, not only performs worse than CCM but is also inferior to some static methods.

\begin{wraptable}{r}{0.3\textwidth} 
    \centering
    \vspace{-15pt}
    \caption{Comparisons among strategies of determining $\boldsymbol{x}_{\mathrm{target}}$, $T_{\mathrm{KDC}}=60$.}
    \label{tab:timesteps_size}
    \resizebox{\linewidth}{!}{%
        \begin{tabular}{l|cl}
            \hline  
            \textbf{Method} & \textbf{$s$} & \textbf{FID} ($\downarrow$) \\
            \hline  
            Single-step & - & 46.82 \\
            \hline
            \multirow{3}{*}{Multi-steps}
            & $0.01$ & 9.96 \\
            & $0.03$ & \textbf{9.32} \\
            & $0.05$ & 9.78 \\
            \hline 
        \end{tabular}
    }
    \vspace{-15pt}
\end{wraptable}
\paragraph{Strategies of determining $\boldsymbol{x}_\mathrm{target}$.}
We tested various methods for determining \( \boldsymbol{x}_{\mathrm{target}} \), including single-step iteration and multiple-steps with different timestep sizes $s$ in Table \ref{tab:timesteps_size}. The effect of directly generating \( \boldsymbol{x}_u \) from \( \boldsymbol{x}_t \) is poor compared to the effect of multi-step generation. This may be because the quality of the directly generated \( \boldsymbol{x}_u \) is relatively low, which affects the effectiveness of CM learning.
We also found that after using CCM, the model is no longer sensitive to timestep sizes, with \( s = 0.03 \) slightly outperforming other choices.

\paragraph{The choice of $T_{\mathrm{KDC}}$.}
\begin{wrapfigure}{r}{0.3\textwidth} 
    \centering
    \vspace{-15pt}
    \includegraphics[width=\linewidth]{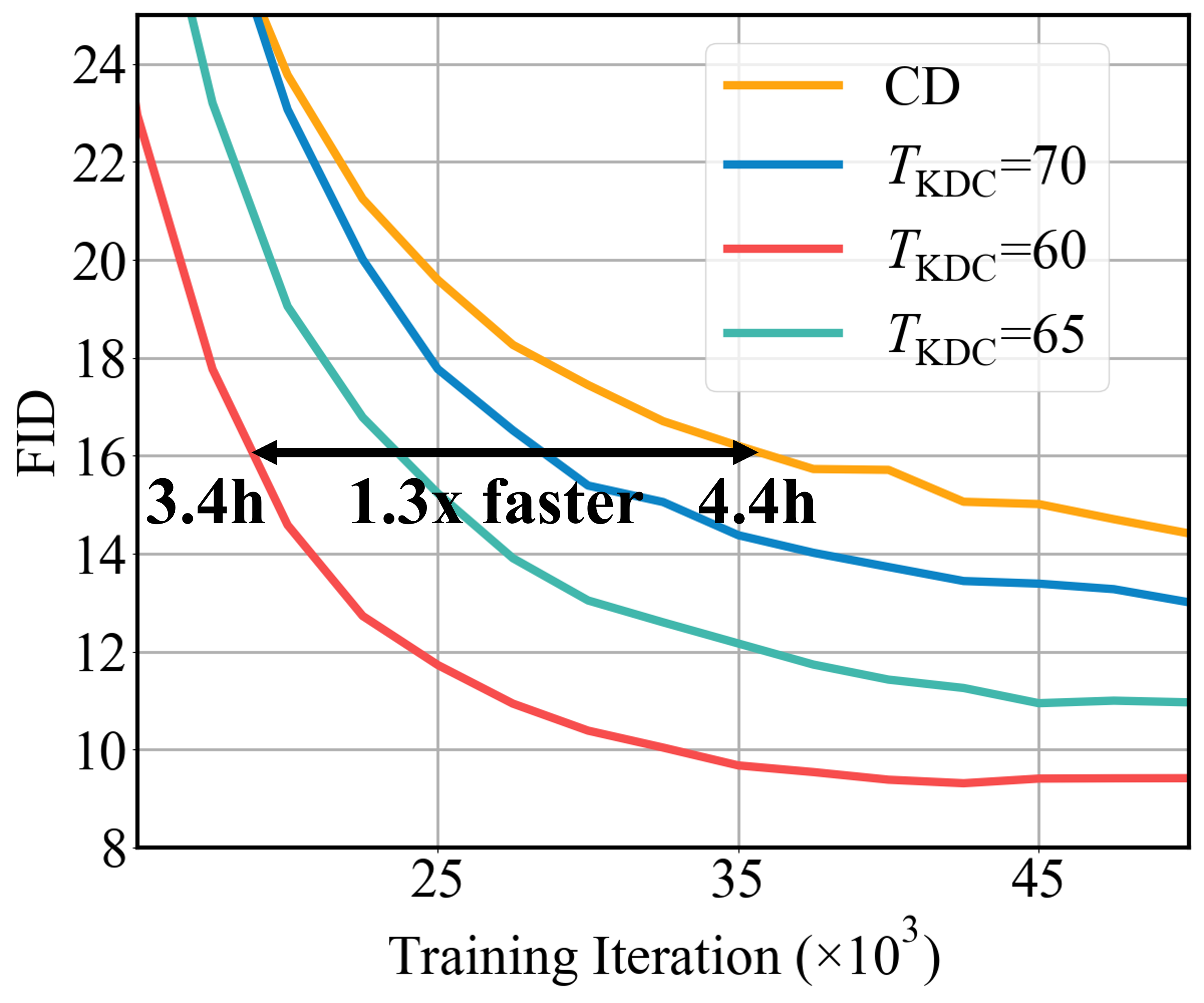}
    \caption{Comparisons of different $T_{\mathrm{KDC}}$.}
    \label{fig:faster}
    \vspace{-15pt}
\end{wrapfigure}
Different $T_{\mathrm{KDC}}$ determine the dynamically selected number of iterative steps during the training process, which is a hyperparameter in the methods presented in this paper.
We conducted experiments with different values of \( T_{\mathrm{KDC}} \), as shown in Figure~\ref{fig:faster}. 
It can be observed that within the range of 60-70, the FID results are better than CD, indicating that our method is not very sensitive to $T_{\mathrm{KDC}}$. Moreover, although CCM will lead to an increase in the time of a single iteration, the convergence rate is accelerated at the same time. Based on the same FID, CCM achieves 1.3× faster convergence than the vanilla CD and achieves a lower FID, bringing significant benefits.
\section{Conclusion}
In this article, we introduce the knowledge discrepancy to measure the difficulty in the CM learning process, and have discovered that the distribution of difficulty is highly imbalanced under different noise intensities. To alleviate this issue, we propose Curriculum Consistency Model (CCM), an efficient method for training models based on ODEs. 
We design an adaptive noise schedule to maintain the consistency of curriculum difficulty and verify the rationality and validity of the design. 
Our method achieves comparable single-step sampling Fréchet Inception Distance (FID) results on CIFAR-10 (1.64) and ImageNet64x64 (2.18). 
More importantly, our approach works on diffusion models and flow matching models as well and we have successfully extended the proposed method to large-scale models, such as Stable Diffusion XL and Stable Diffusion 3. 
We hope that our paper will inspire greater attention to the issue of difficulty in the CM learning process and attract more researchers to engage in related research questions, such as dynamic knowledge discrepancy thresholds, sampling probabilities of \( t \), and so on.

\bibliography{iclr2025_conference}

\begin{thebibliography}{38}
\providecommand{\natexlab}[1]{#1}
\providecommand{\url}[1]{\texttt{#1}}
\expandafter\ifx\csname urlstyle\endcsname\relax
  \providecommand{\doi}[1]{doi: #1}\else
  \providecommand{\doi}{doi: \begingroup \urlstyle{rm}\Url}\fi

\bibitem[Changpinyo et~al.(2021)Changpinyo, Sharma, Ding, and Soricut]{changpinyo2021conceptual}
Soravit Changpinyo, Piyush Sharma, Nan Ding, and Radu Soricut.
\newblock Conceptual 12m: Pushing web-scale image-text pre-training to recognize long-tail visual concepts.
\newblock In \emph{Proceedings of the IEEE/CVF conference on computer vision and pattern recognition}, pp.\  3558--3568, 2021.

\bibitem[Chen et~al.(2018)Chen, Rubanova, Bettencourt, and Duvenaud]{chen2018neural}
Ricky~TQ Chen, Yulia Rubanova, Jesse Bettencourt, and David~K Duvenaud.
\newblock Neural ordinary differential equations.
\newblock \emph{Advances in neural information processing systems}, 31, 2018.

\bibitem[Deng et~al.(2009)Deng, Dong, Socher, Li, Li, and Fei-Fei]{deng2009imagenet}
Jia Deng, Wei Dong, Richard Socher, Li-Jia Li, Kai Li, and Li~Fei-Fei.
\newblock Imagenet: A large-scale hierarchical image database.
\newblock In \emph{2009 IEEE conference on computer vision and pattern recognition}, pp.\  248--255. Ieee, 2009.

\bibitem[Esser et~al.(2021)Esser, Rombach, and Ommer]{esser2021taming}
Patrick Esser, Robin Rombach, and Bjorn Ommer.
\newblock Taming transformers for high-resolution image synthesis.
\newblock In \emph{Proceedings of the IEEE/CVF conference on computer vision and pattern recognition}, pp.\  12873--12883, 2021.

\bibitem[Esser et~al.(2024)Esser, Kulal, Blattmann, Entezari, M{\"u}ller, Saini, Levi, Lorenz, Sauer, Boesel, et~al.]{esser2024scaling}
Patrick Esser, Sumith Kulal, Andreas Blattmann, Rahim Entezari, Jonas M{\"u}ller, Harry Saini, Yam Levi, Dominik Lorenz, Axel Sauer, Frederic Boesel, et~al.
\newblock Scaling rectified flow transformers for high-resolution image synthesis.
\newblock In \emph{Forty-first International Conference on Machine Learning}, 2024.

\bibitem[Geng et~al.(2024)Geng, Pokle, Luo, Lin, and Kolter]{geng2024consistency}
Zhengyang Geng, Ashwini Pokle, William Luo, Justin Lin, and J~Zico Kolter.
\newblock Consistency models made easy.
\newblock \emph{arXiv preprint arXiv:2406.14548}, 2024.

\bibitem[Hang et~al.(2023)Hang, Gu, Li, Bao, Chen, Hu, Geng, and Guo]{hang2023efficient}
Tiankai Hang, Shuyang Gu, Chen Li, Jianmin Bao, Dong Chen, Han Hu, Xin Geng, and Baining Guo.
\newblock Efficient diffusion training via min-snr weighting strategy.
\newblock \emph{arXiv preprint arXiv:2303.09556}, 2023.

\bibitem[Heusel et~al.(2017)Heusel, Ramsauer, Unterthiner, Nessler, and Hochreiter]{heusel2017gans}
Martin Heusel, Hubert Ramsauer, Thomas Unterthiner, Bernhard Nessler, and Sepp Hochreiter.
\newblock Gans trained by a two time-scale update rule converge to a local nash equilibrium.
\newblock \emph{Advances in neural information processing systems}, 30, 2017.

\bibitem[Ho et~al.(2020)Ho, Jain, and Abbeel]{ho2020denoising}
Jonathan Ho, Ajay Jain, and Pieter Abbeel.
\newblock Denoising diffusion probabilistic models.
\newblock \emph{Advances in neural information processing systems}, 33:\penalty0 6840--6851, 2020.

\bibitem[Hoogeboom et~al.(2023)Hoogeboom, Heek, and Salimans]{hoogeboom2023simple}
Emiel Hoogeboom, Jonathan Heek, and Tim Salimans.
\newblock simple diffusion: End-to-end diffusion for high resolution images.
\newblock In \emph{International Conference on Machine Learning}, pp.\  13213--13232. PMLR, 2023.

\bibitem[Hu et~al.(2022)Hu, Wallis, Allen-Zhu, Li, Wang, Wang, Chen, et~al.]{hu2022lora}
Edward~J Hu, Phillip Wallis, Zeyuan Allen-Zhu, Yuanzhi Li, Shean Wang, Lu~Wang, Weizhu Chen, et~al.
\newblock Lora: Low-rank adaptation of large language models.
\newblock In \emph{International Conference on Learning Representations}, 2022.

\bibitem[Huang et~al.(2023)Huang, Sun, Xie, Li, and Liu]{huang2023t2i}
Kaiyi Huang, Kaiyue Sun, Enze Xie, Zhenguo Li, and Xihui Liu.
\newblock T2i-compbench: A comprehensive benchmark for open-world compositional text-to-image generation.
\newblock \emph{Advances in Neural Information Processing Systems}, 36:\penalty0 78723--78747, 2023.

\bibitem[Karras et~al.(2022)Karras, Aittala, Aila, and Laine]{karras2022elucidating}
Tero Karras, Miika Aittala, Timo Aila, and Samuli Laine.
\newblock Elucidating the design space of diffusion-based generative models.
\newblock \emph{Advances in neural information processing systems}, 35:\penalty0 26565--26577, 2022.

\bibitem[Kim et~al.(2023)Kim, Lai, Liao, Murata, Takida, Uesaka, He, Mitsufuji, and Ermon]{kim2023consistency}
Dongjun Kim, Chieh-Hsin Lai, Wei-Hsiang Liao, Naoki Murata, Yuhta Takida, Toshimitsu Uesaka, Yutong He, Yuki Mitsufuji, and Stefano Ermon.
\newblock Consistency trajectory models: Learning probability flow ode trajectory of diffusion.
\newblock \emph{Advances in neural information processing systems}, 2023.

\bibitem[Krizhevsky et~al.(2009)Krizhevsky, Hinton, et~al.]{krizhevsky2009learning}
Alex Krizhevsky, Geoffrey Hinton, et~al.
\newblock Learning multiple layers of features from tiny images.
\newblock 2009.

\bibitem[Lin et~al.(2024)Lin, Wang, and Yang]{lin2024sdxl}
Shanchuan Lin, Anran Wang, and Xiao Yang.
\newblock Sdxl-lightning: Progressive adversarial diffusion distillation.
\newblock \emph{arXiv preprint arXiv:2402.13929}, 2024.

\bibitem[Lin et~al.(2014)Lin, Maire, Belongie, Hays, Perona, Ramanan, Doll{\'a}r, and Zitnick]{lin2014microsoft}
Tsung-Yi Lin, Michael Maire, Serge Belongie, James Hays, Pietro Perona, Deva Ramanan, Piotr Doll{\'a}r, and C~Lawrence Zitnick.
\newblock Microsoft coco: Common objects in context.
\newblock In \emph{Computer Vision--ECCV 2014: 13th European Conference, Zurich, Switzerland, September 6-12, 2014, Proceedings, Part V 13}, pp.\  740--755. Springer, 2014.

\bibitem[Lipman et~al.(2023)Lipman, Chen, Ben-Hamu, Nickel, and Le]{lipman2023flow}
Yaron Lipman, Ricky~TQ Chen, Heli Ben-Hamu, Maximilian Nickel, and Matthew Le.
\newblock Flow matching for generative modeling.
\newblock In \emph{International Conference on Learning Representations}, 2023.

\bibitem[Liu et~al.(2024)Liu, Xie, Deng, Chen, Tang, Fu, Zha, and Lu]{liu2024scott}
Hongjian Liu, Qingsong Xie, Zhijie Deng, Chen Chen, Shixiang Tang, Fueyang Fu, Zheng-jun Zha, and Haonan Lu.
\newblock Scott: Accelerating diffusion models with stochastic consistency distillation.
\newblock \emph{arXiv preprint arXiv:2403.01505}, 2024.

\bibitem[Liu(2022)]{liu2022rectified}
Qiang Liu.
\newblock Rectified flow: A marginal preserving approach to optimal transport.
\newblock \emph{arXiv preprint arXiv:2209.14577}, 2022.

\bibitem[Liu et~al.(2022)Liu, Gong, and Liu]{liu2022flow}
Xingchao Liu, Chengyue Gong, and Qiang Liu.
\newblock Flow straight and fast: Learning to generate and transfer data with rectified flow.
\newblock \emph{arXiv preprint arXiv:2209.03003}, 2022.

\bibitem[Liu et~al.(2023)Liu, Gong, et~al.]{liu2023flow}
Xingchao Liu, Chengyue Gong, et~al.
\newblock Flow straight and fast: Learning to generate and transfer data with rectified flow.
\newblock In \emph{The Eleventh International Conference on Learning Representations}, 2023.

\bibitem[Lu \& Song(2024)Lu and Song]{lu2024simplifying}
Cheng Lu and Yang Song.
\newblock Simplifying, stabilizing and scaling continuous-time consistency models.
\newblock \emph{arXiv preprint arXiv:2410.11081}, 2024.

\bibitem[Lu et~al.(2023)Lu, Lu, Jiang, Szabados, Sun, and Yu]{lu2023cm}
Haoye Lu, Yiwei Lu, Dihong Jiang, Spencer~Ryan Szabados, Sun Sun, and Yaoliang Yu.
\newblock Cm-gan: Stabilizing gan training with consistency models.
\newblock In \emph{ICML 2023 Workshop}, 2023.

\bibitem[Luo et~al.(2023)Luo, Tan, Huang, Li, and Zhao]{luo2023latent}
Simian Luo, Yiqin Tan, Longbo Huang, Jian Li, and Hang Zhao.
\newblock Latent consistency models: Synthesizing high-resolution images with few-step inference.
\newblock \emph{arXiv preprint arXiv:2310.04378}, 2023.

\bibitem[Phung et~al.(2023)Phung, Dao, and Tran]{phung2023wavelet}
Hao Phung, Quan Dao, and Anh Tran.
\newblock Wavelet diffusion models are fast and scalable image generators.
\newblock In \emph{Proceedings of the IEEE/CVF conference on computer vision and pattern recognition}, pp.\  10199--10208, 2023.

\bibitem[Podell et~al.(2024)Podell, English, Lacey, Blattmann, Dockhorn, M{\"u}ller, Penna, and Rombach]{podell2024sdxl}
Dustin Podell, Zion English, Kyle Lacey, Andreas Blattmann, Tim Dockhorn, Jonas M{\"u}ller, Joe Penna, and Robin Rombach.
\newblock Sdxl: Improving latent diffusion models for high-resolution image synthesis.
\newblock In \emph{International Conference on Learning Representations}, 2024.

\bibitem[Pooladian et~al.(2023)Pooladian, Ben-Hamu, Domingo-Enrich, Amos, Lipman, and Chen]{pooladian2023multisample}
Aram-Alexandre Pooladian, Heli Ben-Hamu, Carles Domingo-Enrich, Brandon Amos, Yaron Lipman, and Ricky~TQ Chen.
\newblock Multisample flow matching: Straightening flows with minibatch couplings.
\newblock \emph{arXiv preprint arXiv:2304.14772}, 2023.

\bibitem[Radford et~al.(2021)Radford, Kim, Hallacy, Ramesh, Goh, Agarwal, Sastry, Askell, Mishkin, Clark, et~al.]{radford2021learning}
Alec Radford, Jong~Wook Kim, Chris Hallacy, Aditya Ramesh, Gabriel Goh, Sandhini Agarwal, Girish Sastry, Amanda Askell, Pamela Mishkin, Jack Clark, et~al.
\newblock Learning transferable visual models from natural language supervision.
\newblock In \emph{International conference on machine learning}, pp.\  8748--8763. PMLR, 2021.

\bibitem[Ren et~al.(2024)Ren, Xia, Lu, Zhang, Wu, Xie, Wang, and Xiao]{ren2024hyper}
Yuxi Ren, Xin Xia, Yanzuo Lu, Jiacheng Zhang, Jie Wu, Pan Xie, Xing Wang, and Xuefeng Xiao.
\newblock Hyper-sd: Trajectory segmented consistency model for efficient image synthesis.
\newblock \emph{arXiv preprint arXiv:2404.13686}, 2024.

\bibitem[Rombach et~al.(2022)Rombach, Blattmann, Lorenz, Esser, and Ommer]{rombach2022high}
Robin Rombach, Andreas Blattmann, Dominik Lorenz, Patrick Esser, and Bj{\"o}rn Ommer.
\newblock High-resolution image synthesis with latent diffusion models.
\newblock In \emph{Proceedings of the IEEE/CVF conference on computer vision and pattern recognition}, pp.\  10684--10695, 2022.

\bibitem[Sauer et~al.(2022)Sauer, Schwarz, and Geiger]{sauer2022stylegan}
Axel Sauer, Katja Schwarz, and Andreas Geiger.
\newblock Stylegan-xl: Scaling stylegan to large diverse datasets.
\newblock In \emph{ACM SIGGRAPH 2022 conference proceedings}, pp.\  1--10, 2022.

\bibitem[Song et~al.(2020)Song, Meng, and Ermon]{songdenoising}
Jiaming Song, Chenlin Meng, and Stefano Ermon.
\newblock Denoising diffusion implicit models.
\newblock In \emph{International Conference on Learning Representations}, 2020.

\bibitem[Song \& Dhariwal(2023)Song and Dhariwal]{song2023improved}
Yang Song and Prafulla Dhariwal.
\newblock Improved techniques for training consistency models.
\newblock \emph{arXiv preprint arXiv:2310.14189}, 2023.

\bibitem[Song et~al.(2021)Song, Sohl-Dickstein, Kingma, Kumar, Ermon, and Poole]{songscore}
Yang Song, Jascha Sohl-Dickstein, Diederik~P Kingma, Abhishek Kumar, Stefano Ermon, and Ben Poole.
\newblock Score-based generative modeling through stochastic differential equations.
\newblock In \emph{International Conference on Learning Representations}, 2021.

\bibitem[Song et~al.(2023)Song, Dhariwal, Chen, and Sutskever]{song2023consistency}
Yang Song, Prafulla Dhariwal, Mark Chen, and Ilya Sutskever.
\newblock Consistency models.
\newblock In \emph{International Conference on Machine Learning}, pp.\  32211--32252. PMLR, 2023.

\bibitem[Tong et~al.(2023)Tong, FATRAS, Malkin, Huguet, Zhang, Rector-Brooks, Wolf, and Bengio]{tong2023improving}
Alexander Tong, Kilian FATRAS, Nikolay Malkin, Guillaume Huguet, Yanlei Zhang, Jarrid Rector-Brooks, Guy Wolf, and Yoshua Bengio.
\newblock Improving and generalizing flow-based generative models with minibatch optimal transport.
\newblock \emph{Transactions on Machine Learning Research}, 2023.

\bibitem[Wang et~al.(2024)Wang, Huang, Bergman, Shen, Gao, Lingelbach, Sun, Bian, Song, Liu, Li, and Wang]{pcm2024wang}
Fu-Yun Wang, Zhaoyang Huang, Alexander~William Bergman, Dazhong Shen, Peng Gao, Michael Lingelbach, Keqiang Sun, Weikang Bian, Guanglu Song, Yu~Liu, Hongsheng Li, and Xiaogang Wang.
\newblock Phased consistency model, 2024.

\end{thebibliography}
\bibliographystyle{iclr2025_conference}

\appendix
\clearpage

\section{Experimental Hyperparameters}
\label{app:1}

We minimally change the OT-CFM’s \cite{tong2023improving} design to comply the previous implementation, and important modifications are listed in Table~\ref{tab:hyper}.

\begin{table}[h]
\caption{Experimental details on hyperparameters.}
\label{tab:hyper}
\centering
\resizebox{\linewidth}{!}{
\begin{tabular}{llll}
\hline
\multicolumn{1}{c}{\bf Hyperparameter}  
&\multicolumn{1}{c}{\bf CIFAR-10 32x32} 
&\multicolumn{1}{c}{\bf ImageNet 64x64} 
&\multicolumn{1}{c}{\bf CC3M 1024x1024}
\\
\hline
Training type                       & unconditional & conditional   & conditional   \\
Learning rate                       & 2e-4          & 1e-5          & 5e-6  \\
Discriminator learning rate         & 0.002         & 0.002         & 1e-5  \\
target EMA decay rate $\mu$         & 0.9           & 0.9           & - \\
student EMA decay rate              & 0.9999        & 0.9999        & 0.99 \\
N                                   & 1             & 1             & 1 \\
ODE solver                          & Euler         & Euler         & Euler \\              
Batch size                          & 128           & 2048          & 2     \\
Number of GPUs                      & 1             & 8             & 1     \\
Training iterations                 & 300K          & 500K          & 20k   \\
$T_{\mathrm{KDC}}$                  & 60            & 60            & 60    \\
\hline  \\
\end{tabular}
}
\end{table}
\vspace{-0.5cm}

\section{Additional Experimental Results}
We have conducted experiments to test FID values based on CoCo2014-30K. The results in Table~\ref{tab:30k} demonstrate that CCM performs best on both two types of models.
\begin{table}[H]
  \centering
    \caption{Performance comparisons on CoCo2014-30K}
    \resizebox{0.8\columnwidth}{!}{%
    \begin{tabular}{l|lll}

      \hline
      \textbf{Base Model} & \textbf{Method} & \textbf{CLIP Score} ($\uparrow$) & \textbf{FID} ($\downarrow$) \\
      \hline
      \multirow{4}{*}{SD3} & Original           & 26.18  & 86.84    \\
      & PCM(\cite{pcm2024wang})                 &  31.06    & 28.52      \\
      & LCM(\cite{luo2023latent})               & 31.27  & 25.44     \\
      & \textbf{CCM(ours)}                      &  \textbf{31.41}    & \textbf{21.49}     \\
      \hline
      \multirow{3}{*}{SDXL}
      & Hyper-SD(\cite{ren2024hyper})           &  31.30& 30.87  \\ 
      & PCM(\cite{pcm2024wang})                 &  31.63 & 21.15 \\
      & \textbf{CCM(ours)}                      &  \textbf{31.73} & \textbf{20.47} \\
      \hline
    \end{tabular}
    }
  \label{tab:30k}
\end{table}
\vspace{-0.5cm}

\section{More Samples}
\begin{figure}[h]
    \centering
    \includegraphics[width=\linewidth]{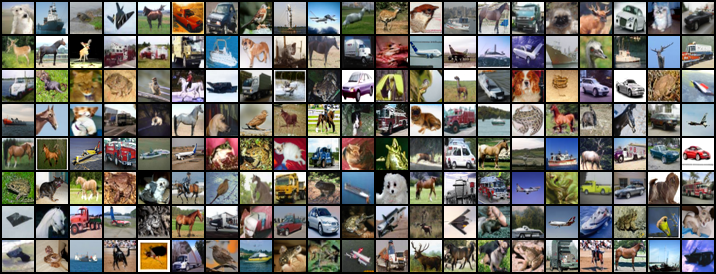}
    \caption{Samples generated by CCM (NFE=1) trained on CIFAR-10.}
    \label{fig:cifar_samples}
\end{figure}

\begin{figure}[h]
    \centering
    \includegraphics[width=\linewidth]{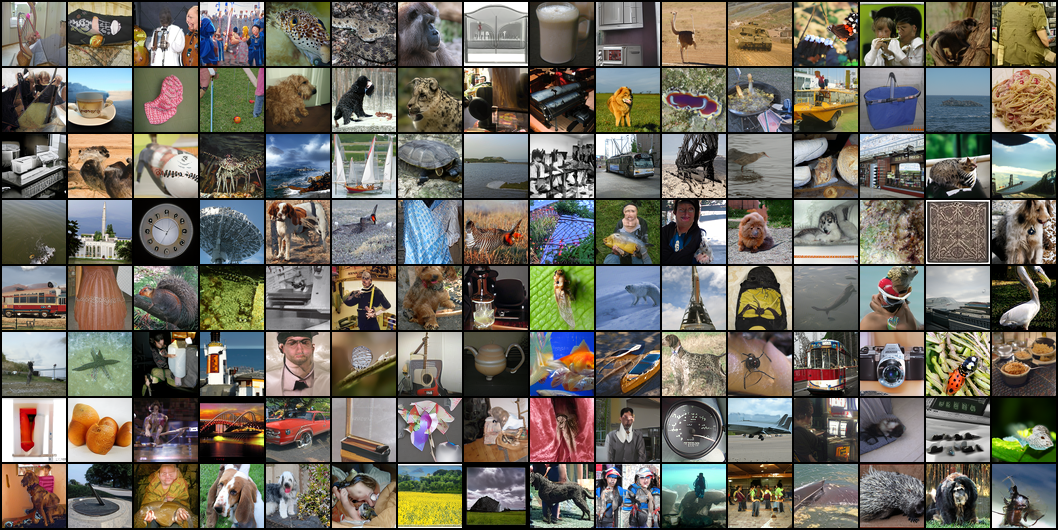}
    \caption{Samples generated by CCM (NFE=1) trained on ImageNet 64x64 according to random classes.}
    \label{fig:imagenet64_samples}
\end{figure}

\begin{figure}[H]
\scriptsize
    \begin{tabular}{@{}m{0.13\linewidth}m{0.82\linewidth}@{}}
    Axolotl & \includegraphics[width=\linewidth]{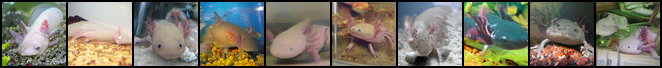} \\ 
    Elephant & \includegraphics[width=\linewidth]{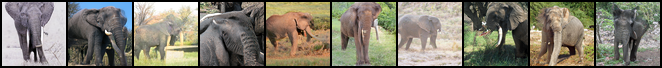} \\ 
    Sheepdog & \includegraphics[width=\linewidth]{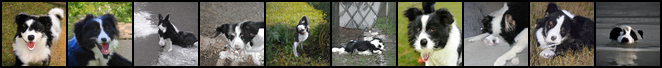} \\ 
    Mud turtle & \includegraphics[width=\linewidth]{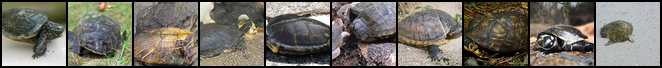} \\ 
    Tree frog & \includegraphics[width=\linewidth]{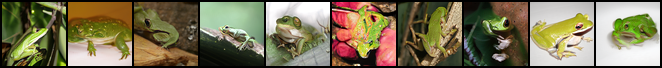} \\ 
    Chickadee & \includegraphics[width=\linewidth]{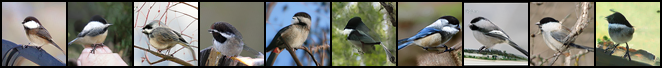} \\
    \end{tabular}
    \caption{Samples generated by CCM (NFE=1) trained on ImageNet 64x64 according to specified classes.}
    \label{fig:cond_samples}
\end{figure}

Here we provide more samples in the Figure~\ref{fig:cifar_samples}-Figure~\ref{fig:cond_samples}.

\end{document}